%% file: main.tex
\documentclass{article}
\usepackage{jfrExamplee}
\usepackage{graphicx}
\usepackage{apalike}
\let\bibhang\relax
\usepackage{natbib}
\usepackage{setspace}
\usepackage{subfiles}
\graphicspath{images/}
\usepackage{amsmath,amssymb}
\usepackage{textcomp,gensymb}
\usepackage{algorithm}
\usepackage[dvipsnames]{xcolor}
\usepackage{hyperref}
\usepackage{epsfig}
\usepackage{epstopdf}
\usepackage{subfig}
\usepackage[noend]{algpseudocode}

\title{Design, Integration, and Field Evaluation of a Robotic Blossom Thinning System for Tree Fruit Crops}

\author{
Uddhav Bhattarai, Qin Zhang, Manoj Karkee\thanks{ Corresponding author: Manoj Karkee, Email: \texttt{manoj.karkee@wsu.edu}} \\
Department of Biological Systems Engineering\\
Center for Precision and Automated Agricultural System \\
Washington State University\\
Prosser, WA 99350 \\
}

\begin{document}

\maketitle

% Instead, growers use assistive platforms like harvest assist systems, with human labor remaining at the core of the operation. Robotics has been limited to controlled environments such as industrial manufacturing sites that require repetitive motion and controlled lighting, which is not the case in tree fruit production.
\begin{abstract}
The US apple industry relies heavily on semi-skilled manual labor force for essential field operations such as training, pruning, blossom and green fruit thinning, and harvesting. Blossom thinning is one of the crucial crop load management practices to achieve desired crop load, fruit quality, and return bloom. While several techniques such as chemical, and mechanical thinning are available for large-scale blossom thinning such approaches often yield unpredictable thinning results and may cause damage the canopy, spurs, and leaf tissue. Hence, growers still depend on laborious, labor intensive and expensive manual hand blossom thinning for desired thinning outcomes. This research presents a robotic solution for blossom thinning in apple orchards using a computer vision system with artificial intelligence, a six degrees of freedom robotic manipulator, and an electrically actuated miniature end-effector for robotic blossom thinning. The integrated robotic system was evaluated in a commercial apple orchard which showed promising results for targeted and selective blossom thinning. Two thinning approaches, center and boundary thinning, were investigated to evaluate the system’s ability to remove varying proportion of flowers from apple flower clusters. During boundary thinning the end effector was actuated around the cluster boundary while center thinning involved end-effector actuation only at the cluster centroid for a fixed duration of 2 seconds. The boundary thinning approach removed 67.2\% of flowers from the targeted clusters with a cycle time of 9.0 seconds per cluster, whereas center thinning approach removed 59.4\% of flowers with the cycle time of 7.2 seconds per cluster. When commercially adopted, the proposed system could help address problems faced by apple growers with current hand, chemical, and mechanical blossom thinning approaches.  

\end{abstract}

% \section{Introduction}

% \section{Related Work}
% \subsection{Machine vision system}

% \begin{figure}[!ht]
% \centering
% \includegraphics[width=0.49\textwidth]{images/ExperimentDetails/FloweringBud.jpg}
% \includegraphics[width=0.49\textwidth]{images/ExperimentDetails/FullBloom.jpeg}

% \caption{Apple flower development within span of 15 days. Each flowering bud can produce around 5 to 7 flowers}
% \label{FolowerDevelopment}

% \end{figure}
\subfile{sections/Introduction}

\subfile{sections/RelatedWork}

\subfile{sections/SystemOverview}

\subfile{sections/Results}

\subfile{sections/Conclusion}

\subsubsection*{Acknowledgments}
This research is partially funded by the Washington Tree Fruit Research Commission (WSTFRC) and United States Department of Agriculture National Institute of Food and Agriculture (USDA NIFA). The authors would like to sincerely thank Ranjan Sapkota, Zixuan He, Achyut Paudel, Martin Churuvija, Safal Kshetri, and Salik Khanal for their support during system integration and field evaluation. Additionally, the support of Priyanka Upadhyaya, Dawood Ahmed, Chenchen Kang, and Atif Bilal Asad during time-sensitive image annotation is highly appreciated. The authors would also like to extend appreciation to Dave Allan (Allan Bros., Inc.) for generously providing access to the orchard during data collection and field evaluation.

\bibliographystyle{apalike}
% \bibliography{jfrExampleRefs}

\end{document}

%% file: sections/Introduction.tex
\section{Introduction}
In 2022, the US apple industry contributed $\sim\$23$ Billion worth of economic activity (direct and downstream) to the US GDP \citep{usapple2022}. However, this important industry (similar to many other tree fruit crops) is largely dependent on semi-skilled human labor for the field operations such as training, pruning, blossom thinning, green fruitlet thinning, and harvesting. Blossom thinning essential in commercial orchards since it balances the number of fruit for better quality and size while ensuring sufficient return bloom next year without compromising the yield this year \citep{robinson2013precision}. If apple flowers are not thinned and all flowers are allowed to set fruit, it will result in a large number of small fruits with reduced fruit size and poor quality. Over cropping also can lead to a depletion of the tree’s carbohydrate reserves, negatively impacting overall tree health. Additionally, many cultivars, such as Fuji and Honeycrisp, suffer from alternate-year fruit set, also known as biennial bearing. To deal with such problems, apple blossom thinning started to be in common practice since the 1940s, which involved the manual removal of flowers \citep{mccormick1933control,bobb1938annual,fioravancco2018biennial}. After more than 80 years, commercial orchards continue to consider blossom thinning as a crucial crop-load management task and perform it annually using various means.

Though manual thinning is a “gold standard”, two mass flower thinning methods, chemical and mechanical thinning, are being used to achieve fast and cost-effective thinning. In chemical thinning, chemicals such as lime and sulfur, ammonium thiosulphate are sprayed when 60-80\% of the flowers are in bloom \citep{mcartney2006effects,myra2007assessment,janoudi2005application,miller2010blossom}. The sprayed chemicals prevent flowers from pollination in different ways including preventing pollen germination and pollen tube growth, burning the flower stigma and pistils, or inhibiting the plant photosynthesis causing plant carbon stress leading to flower abscission. Although chemical thinning can perform mass thinning with reduced cost, the performance of chemical thinning is highly variable and unpredictable depending on the weather, tree age, cultivar, and thinning time (early or late) \cite{robinson2013precision,bound_2014}. In some varieties, chemical thinning might require multiple applications or additional hand thinning to achieve desired crop-load. 

To achieve more control over bloom thinning, mechanical thinners have been developed, which include a rotating center spindle and strings that can be hand-held to cover a section of a canopy such as branch (e.g., Bloom Buster from Automated Ag, WA, USA) or tractor operated to cover a section or entire canopy (e.g. Darwin Thinner from Fruit Tec, Deggenhausertal, Germany). Studies have suggested that Darwin thinner has substantially reduced the thinning cost and thinning time compared to hand thinning in apples, peaches, and plums \citep{darwin_report_canada,reighard2012mechanical}. For example, \citet{schupp2014mechanical} reported that the use of Darwin thinner reduced the thinning time by 27-33\% when compared to the hand thinning alone. However, the current mechanical thinners are not widely adopted by the growers due to their non-selective thinning process and the potential for causing damage to spurs, leaf tissue, and foliage in large canopy regions \citep{dennis2000history,goodfruitgrower_badmecahnicalthinner}. Due to these limitations of the mechanical thinning systems, the majority of the growers still have to rely on chemical thinning for large-scale thinning despite highly variable and uncertain thinning results, and on labor-intensive hand thinning to get the optimal thinning results \citep{goodfruitgrower_handthinning,wideuse_chemical_thinner}. However, apple flowering lasts only a couple of days, depending upon the weather. Hence, a large number of workers are needed over a short period for manual thinning, making the resource and logistics management a challenging task.  Furthermore, it has been widely reported that the labor shortage in agriculture production has been a long-standing problem directly impacting in different agricultural operations including blossom thinning \citep{gallardocovidlaborshortage,agamericalendinglaborshortage,farmlaborshortagehubbart,wallstreetbrat}. Hence it is essential to develop a system that can identify the flower clusters and perform desired intensity of flower removal from the targeted regions while minimally impacting other canopy parts.

This work presents an integrated robotic system to perform targeted, selective blossom thinning that is expected to improve the existing thinning practices in multiple ways. The proposed system combines the mechanical thinning with the intelligence and maneuverability provided by the state-of-the-art machine learning algorithm and robotic manipulation. The integrated system uses a computer vision technique with a RGBD camera and deep learning-based modeling to delineate flower clusters as the locations for targeted thinning. Furthermore, flower cluster position and orientation in 3D space were used to orient the end-effector during the thinning operation. An electrically actuated miniature end-effector, specifically designed for removing flowers within clusters, was designed. The integrated system was tested in the commercial apple orchard, and the performance of the machine vision system, motion planning system, and thinning efficiency have been extensively studied with detailed cycle time analysis.

The remaining parts of the paper are organized as follows. Section \ref{sec_chap6:RelatedWork} presents related work on machine vision and integrated systems for blossom thinning. Section \ref{sec_chap6:SystemOverview} provides an overview of the proposed robotic blossom thinning system. The experimental environment used to test the integrated system is discussed in section \ref{sec_chap6:ExperimentalEnvironment}. . The technical details of the machine vision system, robotic manipulator, and end-effector system are discussed in detail sections \ref{sec_chap6:MachineVisionSystem} and \ref{sec_chap6:MecahtronicSystem}. The field evaluation setup, system hardware, software architecture, and thinning approaches are discussed in section  \ref{sec_chap6:HardwareArchitecture} and \ref{sec_chap6:SoftwareArchitecture}. Results are presented in Section \ref{sec_chap6:Results}, followed by discussion and conclusion in Section \ref{sec_chap6:Discussion} and \ref{sec_chap6:Conclusion}.

%% file: sections/RelatedWork.tex
\section{Related Work} \label{sec_chap6:RelatedWork}
\subsection{Machine vision system for blossom detection}
For a robotic system to perform selective blossom thinning in tree fruit orchards, the system needs to sense the environment and detect and delineate the flowers. Since the flowers are densely located in clusters during full bloom, most of the machine vision systems proposed in the past focused on cluster detection and segmentation. Before the widespread application of deep learning, cluster segmentation was achieved using hand-crafted extraction of color and texture features in RGB images using techniques like intensity thresholding and morphological image processing \citep{krikeb2017evaluation,aggelopoulou2011yield,hovcevar2014flowering}. The performance of feature-based algorithms deteriorates in an unstructured outdoor environment because of the variable lighting condition, uncertainty of the field environment, inherent biological variability of canopies, occlusion of objects (e.g., branches blocking flowers), and variation in object shape, size, color, and other parameters.

In recent years deep learning approaches, specially the convolution neural network (CNN) approaches, are increasingly used for flower and flower cluster detection and segmentation. For semantic segmentation, highly probable cluster regions were first segregated using different CNN-based models such as Clarifai CNN \citep{dias2018apple}, and DeepLab-ResNet \citep{dias2018multispecies,sun2021apple} followed by classification or region delineation techniques such as Support Vector Machine (SVM) \citep{dias2018apple}, Region Growing Refinement (RGR) \citep{dias2018multispecies}, and shape constraint level set \citep{sun2021apple}. The use of fully supervised CNN-based semantic segmentation \citep{wang2020side} and more robust self-supervised learning-based semantic segmentation \citep{siddique2022self} have also been studied. 

In addition to semantic segmentation, cluster detection and instance segmentation of flower clusters have been studied using widely adopted approaches such as Faster R-CNN \citep{farjon2019detection} and Mask R-CNN \citep{bhattarai2020automatic}. Furthermore, researchers have also studied close-range imaging to detect and segment out individual flowers in fruit canopies using channel pruned YOLO \citep{wu2020using} and Mask-Scoring R-CNN \citep{tian2020instance} techniques. The reported deep-learning-based approaches have achieved high accuracy and robustness in flower detection, and thus have shown to be promising techniques for the application in complex, dynamically varying orchard environments. Our prior research (e.g. \citet{bhattarai2020automatic}) had also demonstrated the effectiveness of the deep learning-based method for cluster detection and segmentation. Deep learning based cluster detection and segmentation model was used as the fundamental component of the machine vision system of the proposed robotic blossom thinning system. 

\subsection{Integrated blossom thinning system}
To the best of our knowledge, the reported work is the first study of an integrated robotic system designed and tested for blossom thinning in an outdoor, tree fruit orchard. However, very few studies have been reported that involved using the robotic system for blossom thinning in an indoor environment and integrating a machine vision with an existing thinning machine like Darwin thinner (a mechanical thinning system developed by Fruit Tech, Germany) to improve its intelligence \citep{lyons2015development, ou2012development,lyons_heinemann_2022}. A Specialty Crop Research Initiative (SCRI) project sponsored by USDA developed and tested a prototype robotic system for peach blossom thinning in an indoor environment \citep{ou2012development,pennstatescri}. The prototype system including the in-house designed six degrees of freedom robotic arm was tested for blossom removal using simulated peach branches and blossoms.  The end-effector system included two electrically actuated round brush attachments with an adjustable distance between brush shafts \citep{lyons2015development, ou2012development,lyons_heinemann_2022}. The mechanical system (robotic arm, end-effector, and heuristic programming without the vision system) had 100\% success in removing three artificial blossoms placed 5cm apart in 10, 15, and 20 cm long branches \citep{lyons2015development}. As described by \citet{nielsen2011vision}, the machine vision system included a calibrated camera system with trinocular stereo mounted in an “L” configuration for 3D peach tree construction and peach flower localization. In a separate study \citet{ou2012development} developed manipulator control and path planning algorithm for the manipulator proposed in \citet{lyons2015development} and performed calibration using MATLAB and LabView. The control software took around 7.7 seconds to find the optimal path for the manipulator to reach the target. It was reported that the system needed to be improved for more robust manipulator control, obstacle avoidance, speed, and performance of the machine vision system for experiments in outdoor environments \citep{ou2012development}. In addition to selective thinning with a robotic system, some studies focused on using a stereo or LIDAR-based vision system to adjust the rotation speed \citep{gebbers2013optithin}, position, and orientation \citep{aasted2011autonomous} of Darwin Thinner to improve thinning efficiency based on the flower intensity \citep{gebbers2013optithin} and canopy density \citep{aasted2011autonomous}.

% has been acheieved using single stage detectorand two stage detectors detection using single state and two stage object detectors such as Faster R-CNN (FRCNN), You Only Look Once (YOLO)

% different machine-learning esapproaches are increasingly being used for cluster segmentation. \citet{xiao2014estimation} proposed using a multispectral camera attached in Unmanned Aerial Vehicle (UAV) to estimate flowering intensity. Support Vector Machine (SVM) was implemented and optimized to classify pixels as apple leaves, flowers, and soil based on the R, G, and NIR band information and Normalized Difference Vegetation Index (NDVI). 
% Implementation of deep learning and convolution neural networks are heavily being adopted in semantic segmentation of flower clusters \citep{dias2018apple,dias2018multispecies,}

% \bibliographystyle{apalike}
% \bibliography{jfrExampleRefs}

%% file: sections/SystemOverview.tex
\section{Robotic Blossom Thinning System Overview} \label{sec_chap6:SystemOverview}

A selective blossom thinning system needs to be flexible to offer different thinning strategies depending on the flower intensity, fruit set capability, and crop-load bearing capacity of fruit tree canopies such that the thinning intensity could be tailored to achieve desired crop-load. In pome fruit crops such as apples and pears, flowers are closely spaced in clusters and are often located in constrained working spaces caused by canopy parts such as trunks and branches, and trellis structures such as trellis wires and posts. Once flowers are in full bloom, dense clusters with high flower-to-flower occlusions and larger cluster sizes make individual flower delineation and removal challenging. 

While it is comparatively easier for humans to identify and perform dual arm motion to remove unopened flowers, detection of individual flowers, maneuvering of the robotic arm, estimating precise position and orientation of individual flowers, and positioning of an end-effector is challenging, slow, and computationally intensive. Hence instead of locating and removing individual flowers, the proposed robotic system is developed to remove a proportion of flowers from a cluster of flowers during full bloom. This approach reduces the operation cycle time since the robotic system will have a single target cluster instead of multiple flowers within a cluster. Furthermore, the proportional thinning approach reduces the chances of the robot being in singularity and reduces the computational complexity for the manipulator path planning and obstacle avoidance algorithms. While mechanical blossom thinning systems have demonstrated potential for variable blossom thinning intensities, the major concern from the growers was the damage mechanical thinners could cause to the canopies due to the end effector size. To address the size issue, a miniature electrically actuated end-effector operable within the flower clusters was designed. We propose that the use of robotic system equipped with miniature mechanical thinners with small operational space could be beneficial to the growers, allowing for removal of proportion of flowers en masee while substantially reducing the damage to the spurs and leaf tissue.

\begin{figure}[!ht]
\centering
\includegraphics[scale=0.5]{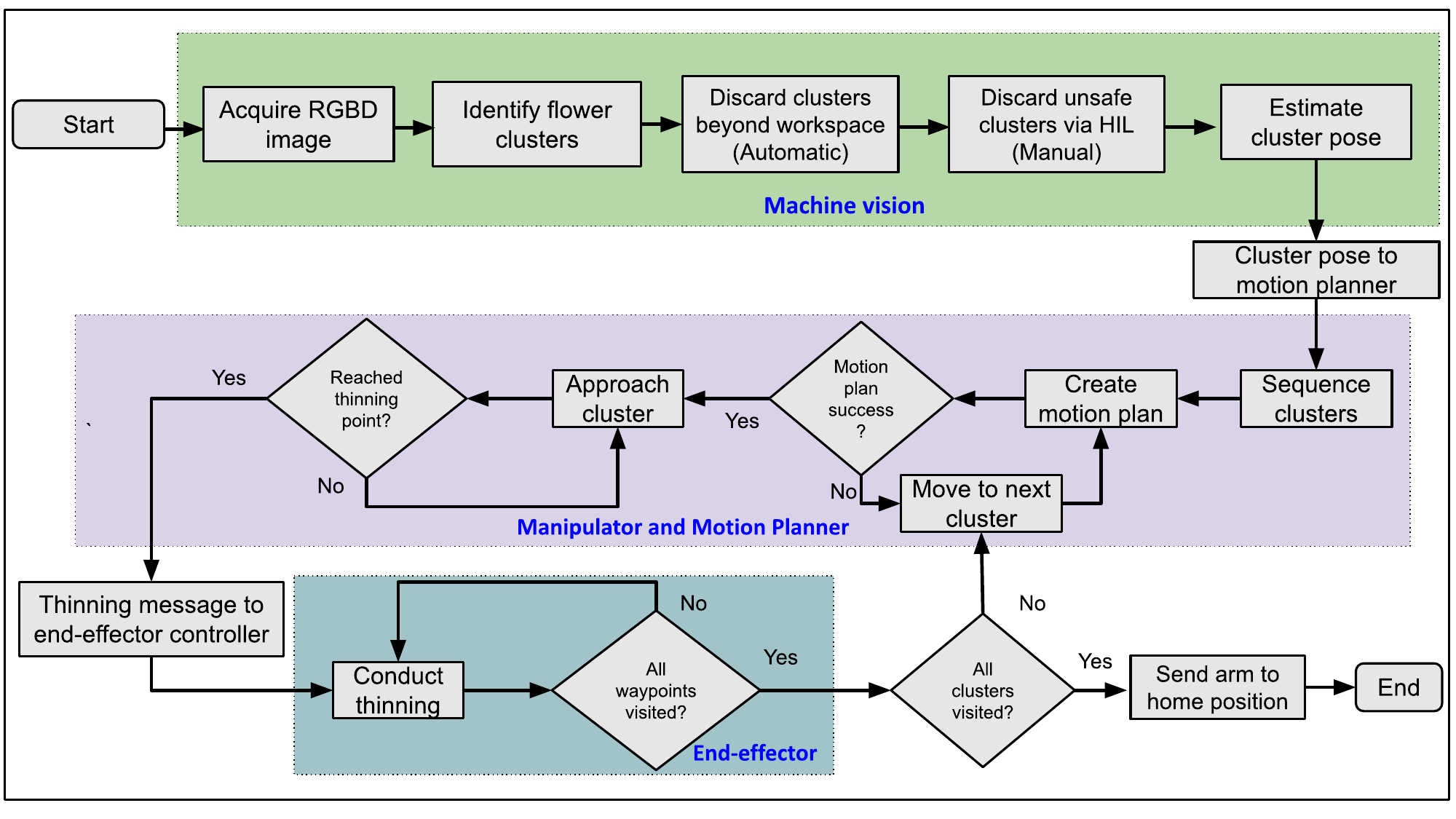}

\caption{Overview of the proposed robotic blossom thinning system. The system consisted of three major components a machine vision system, a manipulator and motion planning system, and an end-effector system. Mask R-CNN based deep learning algorithm was investigated to identify and delineate cluster boundaries followed by cluster pose (position and orientation) estimation. The cluster visit sequence was determined and motion plan for the robotic manipulator was developed to navigate end effector to desired thinning position and orientation to conduct thinning.}
\label{fig_chap6:System Overview}
\end{figure}

Figure \ref{fig_chap6:System Overview} illustrates the overview of the proposed robotic blossom thinning system consisting of three major subsystems: machine vision system, robotic manipulator and motion planning system, and an end-effector system. The machine vision system included associated sensors and an image processing pipeline for sensing, identifying, and delineating flower clusters. The identified flower clusters were examined for safe reachability using automated and manual (human in loop) approach. Clusters beyond 1m depth, and the clusters with invalid depth measurements due to camera sensor error were automatically excluded. The remaining clusters were further manually evaluated to ensure the safe reachability of the manipulator for any collisions with other orchard components, including the trunk, branches, trellis wires, and training posts.  
The flower clusters in orchards were randomly distributed with random position and orientation. Hence, the cluster pose (position and orientation) was estimated such that the end effector could be positioned depending upon the cluster orientation to achieve desire thinning result. In this work the end effector was positioned orthogonal to the cluster surface during the thinning operation. Based on the cluster pose information from the machine vision system, visit to the individual clusters were prioritized by computing shortest path to be followed by the manipulator such that each clusters were visited once.  Once the cluster visit was prioritized, the corresponding motion plan for six degrees of freedom robotic manipulator was computed and executed to navigate the end effector to the desired position and orientation to perform thinning. The end-effector system actuated the end-effector until all the thinning waypoints were visited depending upon the thinning strategy. The thinning waypoints were defined depending on the employed thinning strategies. Two thinning strategies (Boundary Thinning and Center Thinning) were studied to demonstrate the ability to control thinning intensity to perform selective thinning from target clusters. For boundary thinning the end effector was navigated and actuated around the cluster boundaries while for center thinning, the end effector was navigated to the cluster centroid and actuated for fixed 2 seconds. The following sections discuss the integrated system’s experimental environment and various hardware and software subsystems in more detail.

\section{Experimental Environment} \label{sec_chap6:ExperimentalEnvironment}
\begin{figure}[h!]
\centering
\includegraphics[width=0.48\textwidth]{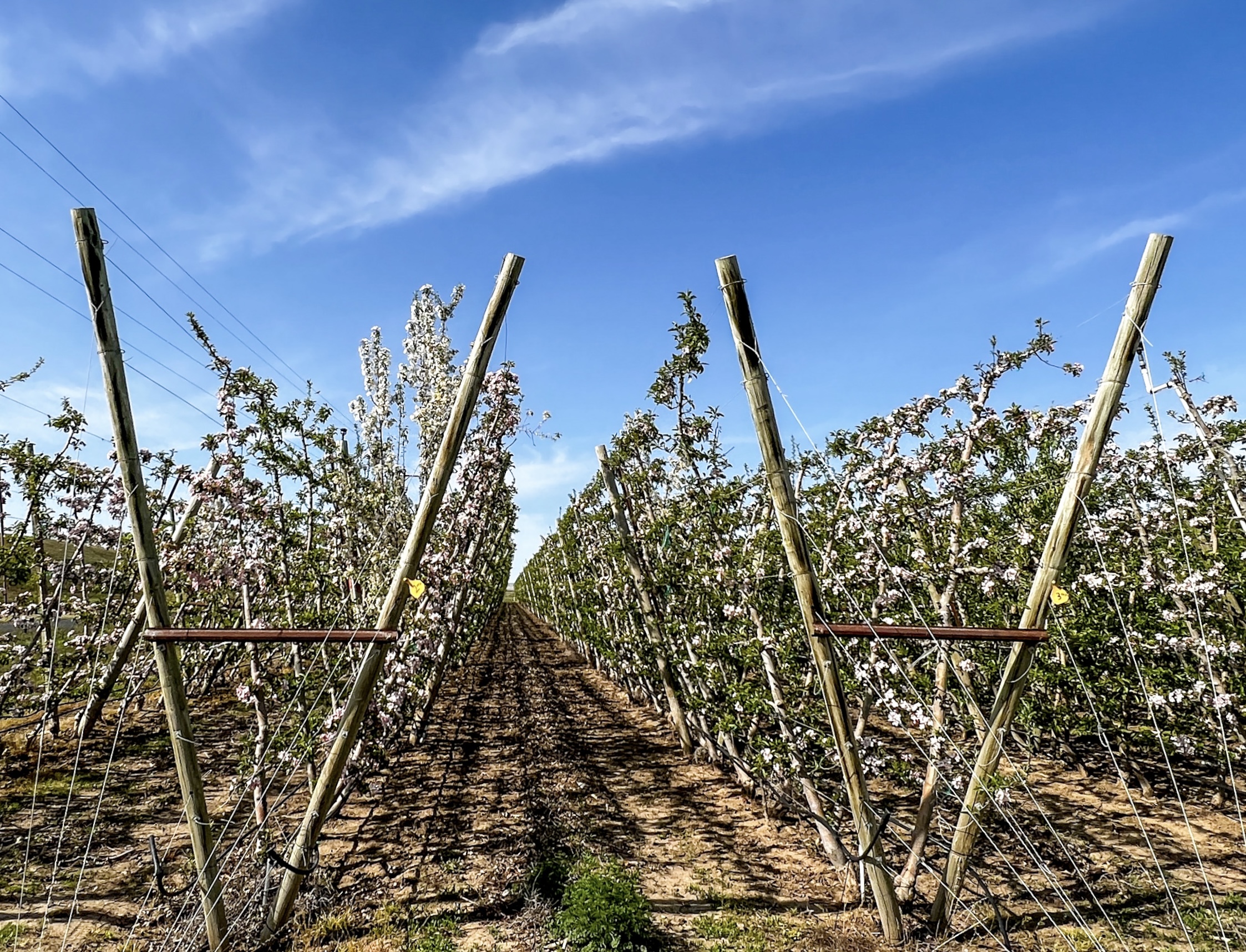}
\includegraphics[width=0.48\textwidth]{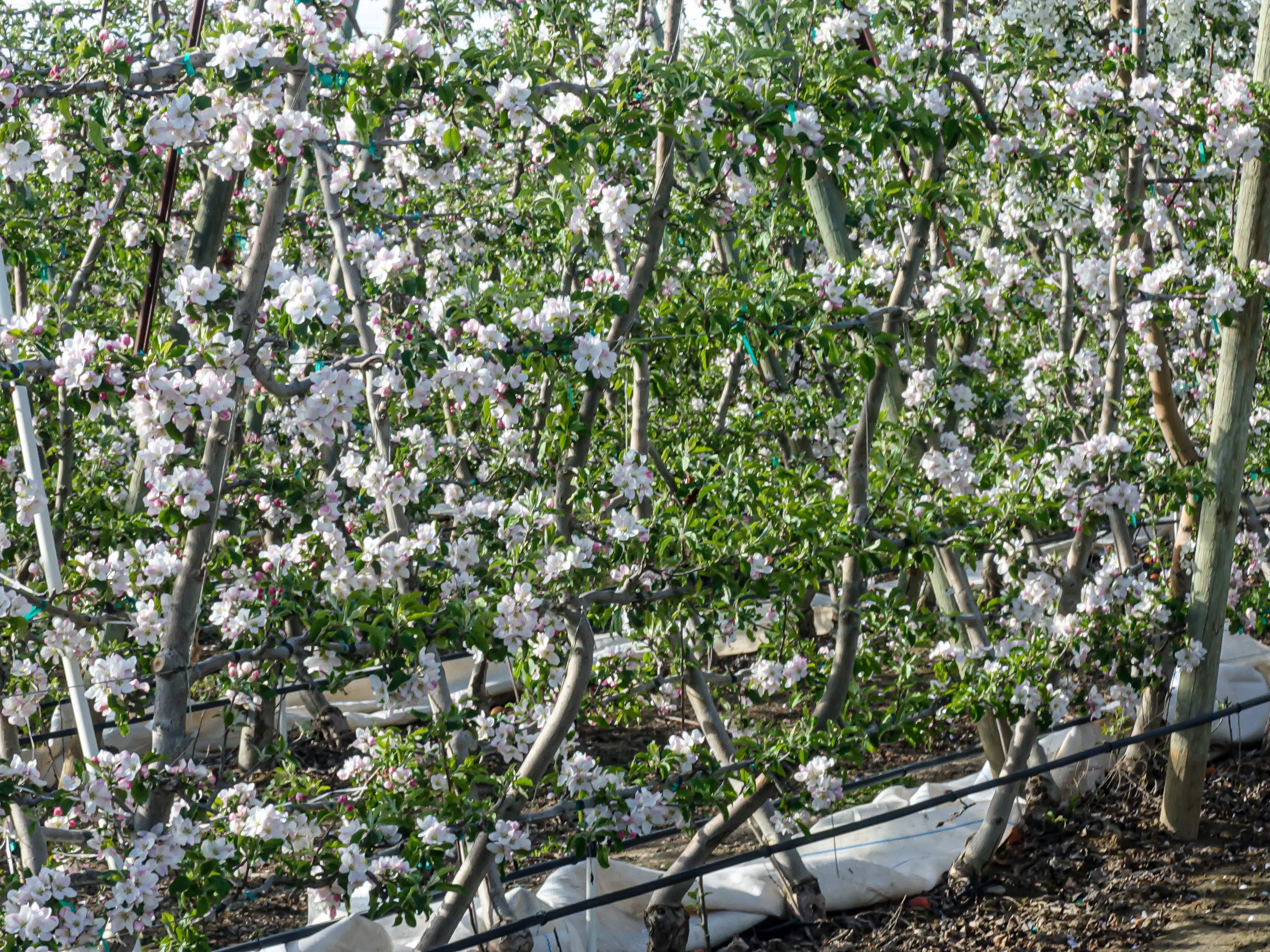}

\caption{The experimental orchard (Envy apple variety) used for evaluating the robotic blossom thinning system. The trees in this experimental site were trellis-trained to create a V-shaped fruiting wall orchard.}
\label{chap6EnvyOrchard}

\end{figure}

A commercial apple orchard growing Envy (\textit{Scilate} cultivar) apple variety in Prosser, WA was selected for field test of the proposed robotic flower thinning system. The apple trees were grown in a modern trellis-trained high-density planar architecture as shown in Figure\ref{chap6EnvyOrchard}. The trees were trained to a V-trellised system leading to an inverted pyramidal structure of the canopies. Traditionally, apple trees are planted in rows with wide spacing and trunks and branches are allowed to grow in all directions, which leads to a three-dimensional canopy structure resulting in 4D working space, making it highly challenging for any robotic operations. The modern orchard systems use trellis wires to support and train fruit trees to create a narrow vertical plane of canopy with flower and fruit growing mostly in the outer surface (also called a fruiting wall architecture). This simplified working space helps improve accessibility, efficiency and productivity of both human workers as well as mechanical and robotic machines. 

\begin{table}[!ht]
    \centering
    \caption{Attributes of the V-trellised orchard used for field evaluation of the robotic thinning system (from \citet{davidson2016hand})}
    \label{tab_chap6:EnvySpecifications}
    \setlength{\tabcolsep}{3pt}
    
    \begin{tabular}{lccccccc}
        \hline \hline
         & Variety &Canopy Angle &No. of Trellis &Trellis Wire Spacing &Tree Height &Tree Spacing &Row Width\\
         
         & &(Degree) &Wires &(cm) &(cm) &(cm) &(cm) \\
         \hline
         & Envy &75 &7 &46$\pm$1 &366 &142$\pm$19 &353$\pm$3 \\
         \hline \hline
    \end{tabular}

\end{table}

\section{Machine Vision System} \label{sec_chap6:MachineVisionSystem}
A machine vision system is critical for any robotic system to operate. The vision system provides a capability to robots to locate objects it would like to manipulate (apple flowers in this case), the objects that need be avoided (e.g., trunk, branch, trellis wire, support posts), and the other components (leaves) in the robot’s workspace. As a first step in developing an autonomous robot for blossom thinning, this work focused on developing deep learning-based models to detect, segment, and localize apple flower clusters in images collected from outdoor orchard environment. 

\subsection{Eye-in-hand camera system}
The machine vision system developed for this work used an Intel RealSense D435i RGB-D camera as eye in hand sensor that was rigidly attached to the tool flange of the robotic manipulator. Such attachment allowed fixed transformation between the camera to the robot coordinate space while providing multi-view imaging of objects when the robot is in motion. The camera system used a single RGB imaging sensor capable of providing a maximum resolution of $1920 \times 1080$ pixels for the given Field of View (FOV) of  $69\degree  \times 42\degree$. Depth was estimated using active stereo IR technology with a maximum resolution of  $1280 \times 720$ with FOV of $87\degree \times 58\degree$. In this work, the RGB and depth image resolution of $1280 \times 720$ pixels was used to speed up the object segmentation pipeline while covering the region within the reachability of the manipulator.

\subsection{Flower cluster detection and segmentation}
Since the integrated system requires the manipulator to navigate the end-effector that can engage with the individual clusters in a unique way, each segmented cluster requires a unique identifier. This work used the Mask R-CNN model \citet{he2017mask} for flower cluster detection and pixel-level instance segmentation. The instance segmentation technique was selected over semantic segmentation since instance segmentation separates the multiple instances of the same object class with unique identifiers. Furthermore, the end-effector had to interact within the cluster more precisely (e.g., following the flower cluster boundary); a pixel-level segmentation algorithm was well suited instead of bounding box-based object detection methods such as Faster R-CNN and You Only Look Once (YOLO).

In our previous work, the Mask R-CNN model showed promising accuracy and robustness in detecting and segmenting flower clusters in outdoor orchard images \citep{bhattarai2020automatic}.  However, when the same model was applied to the images acquired from the eye-in-hand camera attached to the robotic manipulator, the model struggled to identify clusters. One of the major reasons for the decreased cluster detection was due to the closer imaging distance $(\sim50cm - \sim 75cm)$ for the eye-in-hand camera, which differed from the $\sim 1m$ distance used in the previous work \citep{bhattarai2020automatic}. This was verified by manually moving the camera away from the canopy which resulted in improved cluster detection. Hence to train the Mask R-CNN model for close range images, new image dataset was collected from a commercial apple orchard in Prosser, WA, using Intel RealSense D435i RGB-D camera at the resolution of $1280 \times 720$ for both color and depth images. Tree canopies were continuously scanned at a distance of $\sim 70 cm$ (similar to distance of the eye in hand camera) from the trunk in a daylight, and the images frames were extracted later from the recorded data stream. The flower clusters in these images were annotated as polygons using the Labelbox image annotation tool (https://labelbox.com/). The image dataset consisted of 636 images with a total of 2,118 flower clusters, which were randomly divided into training, validation, and test sets with a split ratio of $70\% : 15\% : 15\%$ images. 

A Mask R-CNN model was then trained extending the existing Mask R-CNN implementation from Matterport Inc.(Sunnyvale, CA) released under MIT License \citep{matterport_maskrcnn_2017}. The system employed TensorFlow 2.0 with Keras as backend engine and ResNet101 as the backbone. The ResNet101 backbone was pre-trained with the
MSCOCO dataset. The network was trained up to 100 epochs using Stochastic Gradient Descent (SGD) search technique with a momentum of 0.9 and a Learning Rate (LR) of 0.00005. Different image augmentation techniques, including random horizontal and vertical flips, random scaling ($0.5$ -- $1.5$), and random rotation ($-60 \degree$ -- $+60\degree$), were used to increase the dataset variability.

The segmented cluster boundary consisted of a large number of boundary points as the Mask R-CNN model was trying to closely and smoothly locate the cluster boundary. However, to perform robotic thinning, it was not necessary and practical for the end-effector to navigate to all the vertices of the polygon representing the flower cluster boundaries. First, navigating to all the vantage points (vertices) would result in additional computational overhead, which is complex and time-consuming for the motion planner. Second, in most cases, the end-effector was large enough to cover multiple vertices on the boundary at once, making individual visits to vertices redundant. Hence, a Convex Hull algorithm was used to simplify the cluster boundary polygons. The objective of the Convex Hull algorithm was to estimate the smallest convex polygon such that all the cluster boundary points were either on or within the polygon representing the cluster boundary. The implemented Convex Hull algorithm was available in the OpenCV library and had $O(NLogN)$ complexity.

\subsubsection{Flower Cluster Reachability}
Once the potential flower clusters were segmented, individual clusters were evaluated for potential to reach them safely (without any robot-canopy collision). The flower clusters that did not satisfy the robot reachability requirements were filtered out from the target clusters for thinning. First, clusters beyond 1m depth and having invalid depth due to camera sensor error were first filtered out automatically. The remaining clusters were further evaluated for reachability by the system’s operator as a part of the human-in-the-loop system. The objective of the human in the loop system was to examine the detected clusters from the perspective of proximity to the trunks, branches, and trellis wires, and remove clusters from the cluster target list if navigation to, and thinning of those clusters could cause potential collision between the robotic system and the canopy parts. The effort of the current human-in-loop system could be reduced by detecting and segmenting other canopy parts such as trellis wires, trunks, branches, leaves, and support posts, and making automated thinning decisions based on the operation environment. The current deep learning-based machine vision system developed for cluster segmentation is flexible and can be extended to identify other canopy parts with few modifications.

% Although current training systems are more accessible for robots to operate, there is still a challenge for robots is complete field operations without human intervention. Because of the permanent trunk structure, branches, trellis wires, leaves, support posts, and environmental factors such as variable lighting conditions, it is challenging for robots to complete field operations on their own.

\subsection{Cluster Localization}
Flower clusters grow in random/variable locations and orientation within constrained canopy spaces. It is critical to determine the precise position and orientation of the clusters in 3D space such that the thinning end-effector can be positioned and oriented precisely for effective thinning. In this work, the cluster position was estimated by projecting the detected cluster pixels in 3D space and averaging the corresponding 3D cluster position. The cluster orientation was then computed by estimating the segmented cluster’s surface normal vector as illustrated in Algorithm \ref{chap6cluster_orientation_estimation}. Let $P_1 \dots P_N$ be the 3D position of all points in the flower cluster with camera viewpoint location $(v_p)$. To estimate the surface normal of the entire cluster, the cluster was divided into sub-sections defined by radius $(R)$ and number of neighborhood points $(k)$. The surface normal of each cluster subsection was estimated and then averaged to compute the surface normal of the entire cluster (see Algorithm \ref{chap6cluster_orientation_estimation}).  

\begin{algorithm}[!h]
\caption{Flower cluster orientation estimation}
\begin{algorithmic}[1]

% \Procedure{EstimateOrientation}{$a,b$}       \Comment{This is a test}
    \Require{$P_1 \dots P_N$}, camera viewpoint $(v_p)$                
    \Ensure{Cluster position ($\textbf{P}$), Cluster normal ($\vec{\textbf{n}}$)}
    \Statex
    \Function{EstimatePositionOrientation}{$C[\;]$}
        \State $\textbf{P}=\frac{1}{N}\sum_{i=1}^N {P_i}$
        
        \State{$R$ $\gets$ $0.1$}
        \State{$k$ $\gets$ $30$}
        \State{Normal vector list $n_l$ $\gets$ $[\;]$}
        
        \For{$i \gets 1$ to $N$}
        \State{Build a $k-D$ tree}
        \State{Conduct radius $(R)$ neighbourhood search for $k$ neighbours}
        \State Estimate covariance matrix $C$
        \begin{equation}
            C=\frac{1}{k}\sum_{l=1}^k {(p_l-\overline{p})}\cdot {(p_l-\overline{p})^T},\overline{p}=\frac{1}{k}\sum_{l=1}^k{p_l}, p_l\in P^k
        \end{equation}
        % \begin{equation}
        %     Where, \overline{p}=\frac{1}{k}\sum_{i=1}^k{p_i}, p_i\in P^k
        % \end{equation}
        \State Estimate eigenvalues of $C$:  $C\cdot \vec {v_j}=\lambda\cdot \vec {v_j}, \textit{j}\in {\{0,1,2\}}$, where, $0\leq\lambda_0\leq\lambda_1\leq\lambda_2$
        
        \State Surface normal $\vec{n_i}={\{n_x,n_y,n_z\}}$; vector $\Vec{v_0}$

        \State Orient $\vec{n_i} \ni \vec{n_i} \cdot (v_p-p_i)>0$
        \State Append $\vec{n_i}$ to $n_l$
        \EndFor
        \State Cluster normal $\vec{\textbf{n}}=\frac{1}{m}\sum_{i=1}^m{\vec{n_i}}, \vec{n_i}\in n_l$

        \State \Return {$\textbf{P}, \vec{\textbf{n}}$}
        \EndFunction

\end{algorithmic}
\label{chap6cluster_orientation_estimation}
\end{algorithm}

\section{Mechatronic System} \label{sec_chap6:MecahtronicSystem}
\subsection{Robotic Manipulator and Motion Planning}
For the given position and orientation of the flower cluster, the objective of the manipulator was to navigate the end-effector to the thinning location orthogonal to the cluster surface to conduct blossom thinning. An UR5e collaborative robotic manipulator manufactured by Universal Robots (Odense, Denmark) was used in this study. The manipulator has a small footprint, a 5KG payload that was sufficient for orchard operations such as thinning, pruning, and harvesting, and safety functions related to joint speed, torque limit, Tool Center Point (TCP) pose, speed, and force limit, among others. Table \ref{tab_chap6:UR5eSpecifications} details the robotic manipulator specifications. The experimental orchard used was trained to 2D planar structures making a large proportion of flowers accessible for the robotic thinning. The robot motion planning was achieved using the default ROS MoveIt Open Motion Planning Library (OMPL).

%Hence, in the current situation collaboration between humans and robots is likely such that productivity is improved. Hence it is desirable to have robotic systems that can collaboratively work with humans. It is desirable to have robots with flexibility and multitasking, lower footprint, and enough payload and productivity while maintaining safety for the robot, orchard, and humans. This provides huge potential for collaborative robots to operate in the field. Collaborative robots are designed to operate with humans and help humans in certain aspects of problem-solving. 
\begin{table}[!ht]
    \centering
    \caption{UR5e robot technical specifications \citep{UR5specs}}
    \label{tab_chap6:UR5eSpecifications}
    \setlength{\tabcolsep}{3pt}
    \scalebox{1}{
    \begin{tabular}{lccccccc}
        \hline \hline
         &DOF &Payload (kg) &Reach (mm) &Footprint $\phi$ (mm) &Weight (kg) &Power Consumption (W)\\
         \hline
         &6 & 5 &850 &149 &20.5 &200\\
         \hline \hline
    \end{tabular}
    }

\end{table}

\subsection{End-effector design}
The major requirement of the blossom thinning end-effector was to exert a high impact on the target flowers to be removed while minimally affecting the flowers to be saved for pollination as well as neighboring leaves and spurs. Flowers are tightly located in clusters, often overlapping each other. Hence, the end-effector should also be small and precise, impacting only at the desired location in flower clusters. As a preliminary experiment, three different actuation mechanisms were considered: water-actuated, air-actuated, and electrically-actuated end-effectors. For both the water and air actuated system, water and air were pressurized in the high-pressure tank and were released as a water jet/air stream using a sub-millimeter size nozzle. Although both water jet and air stream approaches removed some of the flowers, two major problems were observed with these thinning approaches; i) While removing the unnecessary flowers, the flowers that should have been saved were frequently impacted/damaged by the incoming water jet and air stream often dragging them with the water/air flow; ii) The incoming water jet or air stream also induced damage to the flowers in proximal clusters and the clusters in the background. After the preliminary experiments, waterjet and air stream techniques were, therefore, dropped and only the electrically actuated brushing mechanism was further investigated.
\begin{figure}[!h]
\centering
\includegraphics[scale=0.08]{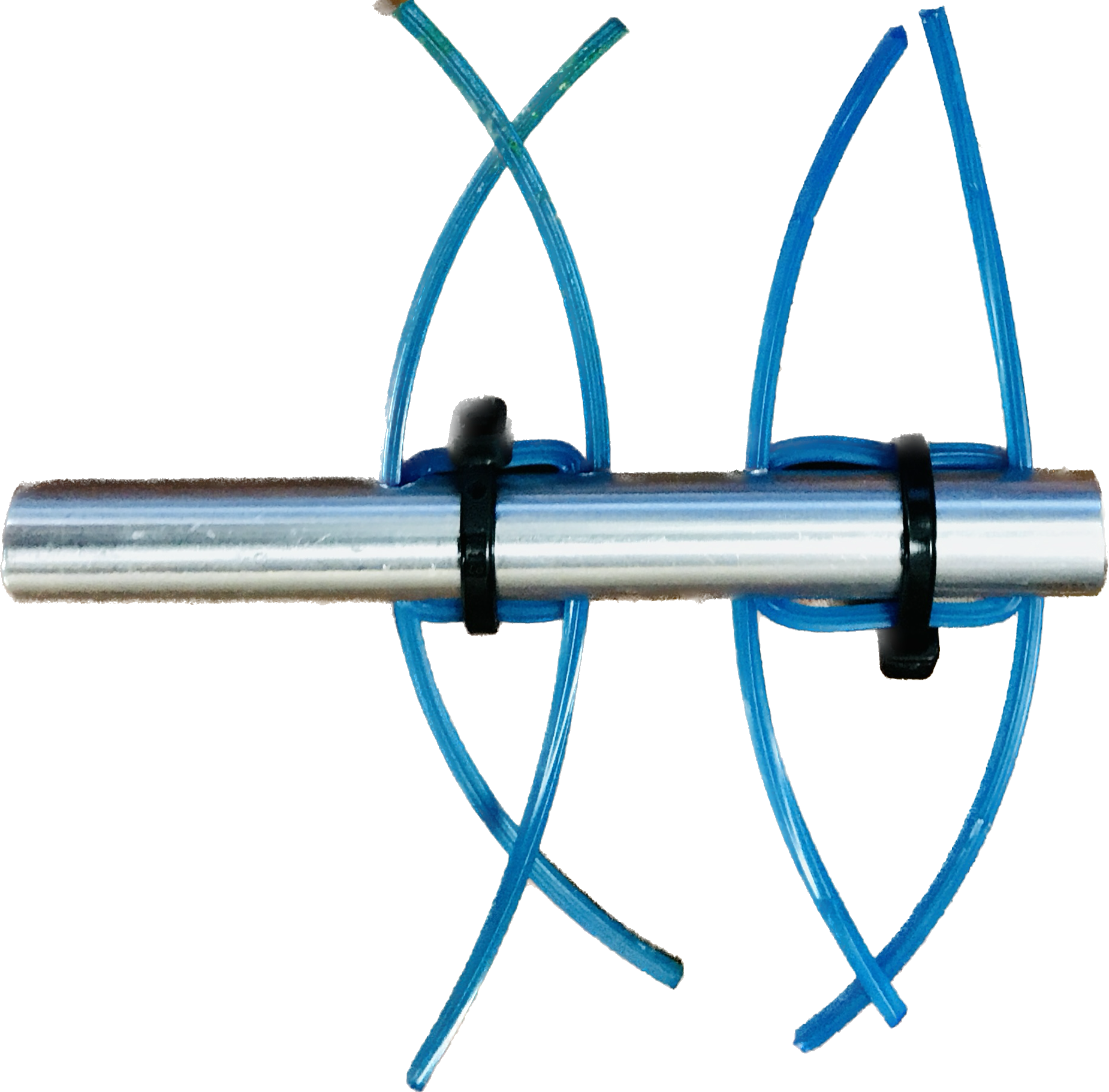}

\caption{Electrically actuated end-effector design with spindle string. Strings used were the pieces of off-the-selves grass trimming lines with grooved edges.}
\label{fig_chap6:endeffector}
\end{figure}

The electrically actuated end-effector system consisted of a miniature spindle string configuration. Commercial-grade string trimmer lines were tied together in an aluminum spindle at $\sim 7mm$ spacing, as shown in Figure \ref{fig_chap6:endeffector}. The strings have grooved edges for faster trimming. This end-effector was coupled with a high-speed motor with a rigid shaft coupler to provide the necessary force to thin flowers quickly and efficiently. During the preliminary experiment, the end-effector was held manually to mimic robotic thinning motion—a clockwise sweeping motion around the cluster boundary with the approach direction orthogonal to the cluster surface. Three different experiments were conducted by varying the motor speed: i) Low speed (3000rpm-4000rpm), ii) Medium speed (5000rpm-6000rpm), and iii) High speed (7000rpm-8000rpm). For each thinning speed, 30 flower clusters were thinned to evaluate thinning efficiency. High-speed motor actuation showed better thinning efficiency compared to others with faster trimming of excess flowers while impacting the neighboring flowers in clusters minimally. Furthermore, it was observed that the trimmer line strings were rigid enough to remove flowers, and minimally damaged branches and spurs in small areas in the event they were present.

\section{Experimental Setup for Field Evaluation} \label{sec_chap6:HardwareArchitecture}
\begin{figure}[!h]
\centering
\includegraphics[scale=0.45]{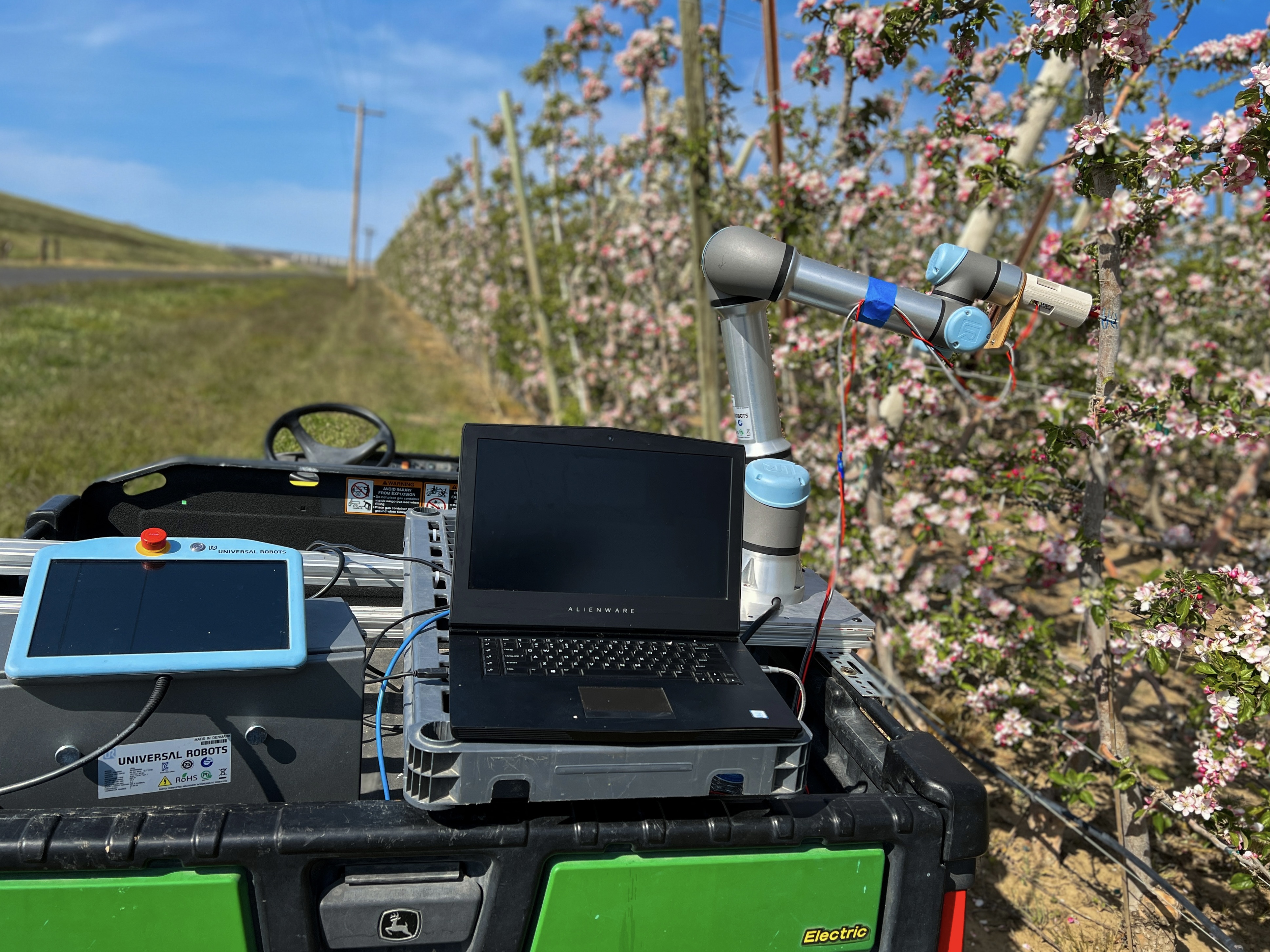}

\caption{Integrated system during the field evaluation in V-trellised fruiting wall architecture commercial apple orchard in Prosser, WA. The UR5e robotic manipulator was attached rigidly at a back of a utility vehicle.}
\label{fig_chap6:IntegratedSysteminField}
\end{figure}

Figure \ref{fig_chap6:IntegratedSysteminField} shows the experimental setup during the field test in a commercial apple orchard in Prosser, WA. All the hardware components were mounted on an electric utility vehicle (John Deere, IL, USA). The mobile platform was driven along the apple tree rows during the test. The experiments were conducted in stop, execute, and go fashion, where the mobile platform was stationary when the manipulator and end-effector were actuated for blossom thinning. The weather condition during the field test was sunny to cloudy, with temperatures ranging from $10-25\degree C$. The wind was variable, ranging from minimal (5mph) to high (18mph). 

\begin{figure}[!ht]
\centering
\includegraphics[width=\textwidth]{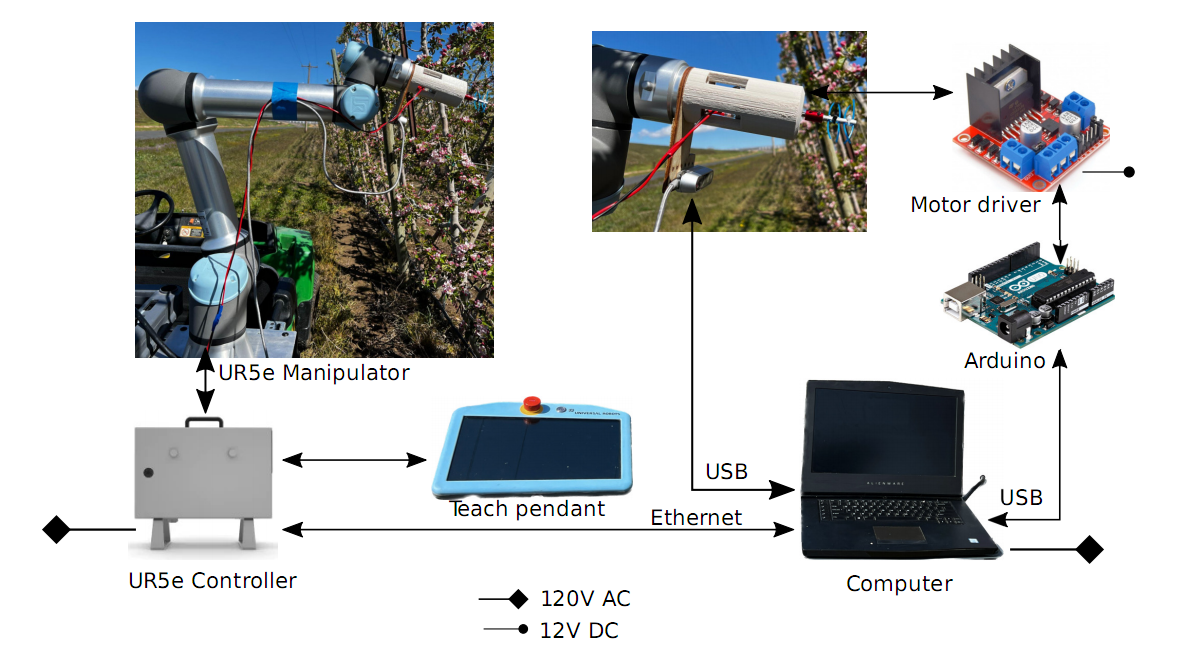}
\caption{Hardware components used during field test and communication protocol. Dell Alienware15R4 laptop was used as the central processing system of the integrated system.  The laptop communicated with the Intel RealSense camera and Arduino via USB connection while LAN connection was established between the UR5e controller and the laptop to control the UR5e manipulator.}
\label{fig_chap6:IntegratedSysteminCommunication}
\end{figure}

Figure \ref{fig_chap6:IntegratedSysteminCommunication} shows the hardware components of the robotic system, which primarily included a 6DOF – UR5e manipulator, UR5e controller, UR5e teach pendant, laptop, Intel RealSense 435i RGB-D camera, H-bridge motor driver and Arduino Uno, DC 777 motor, and an end-effector. The manipulator and the laptop were electrically powered by Honda EU2200i (2200 W, 120 V AC) portable generator (Honda Motor Co. Ltd, Tokyo, Japan), while a 12V DC battery powered the end-effector motor. The laptop used was Dell Alienware15R4, with Intel Core i7-8750 CPU, 2.2 GHz processor, 64-bit operating system, 32GB RAM, and 8GB NVIDIA GeForce GTX 1080 Graphics Processing Unit (GPU) running Ubuntu 18.04 LTS operating system. The laptop interacted with the manipulator via Local Area Network (LAN) connection via the UR5e controller box. Communication between the laptop and Intel RealSense camera and Arduino microcontroller was established via a USB 3.0 connection. Since Arduino could only supply up to a 5V signal, the H-bridge motor driver was used as a bridge to control the 12V supply from the battery based on the signal from Arduino.

\subsection{Thinning approaches}

As discussed in Section \ref{sec_chap6:SystemOverview}, selective thinning requires the ability to customize the thinning intensity based on different factors such as flower density, fruit set capability, cultivar, and size and distribution of branches and trunks of target fruit trees. To demonstrate the system’s ability to perform variable thinning, two thinning strategies were investigated for flower removal. 
\begin{itemize}
    \item Boundary thinning: In this approach, size of individual clusters was estimated, and the end-effector was actuated around the cluster boundary in a clockwise sweeping motion. 
    \item Center Thinning: In center thinning approach, the end-effecter was moved to the cluster centroid and rotated for a specific duration (2 sec) irrespective of the cluster size and without sweeping motion used in the boundary thinning.  
\end{itemize}

\begin{figure}
    \centering
    \includegraphics[scale=0.3]{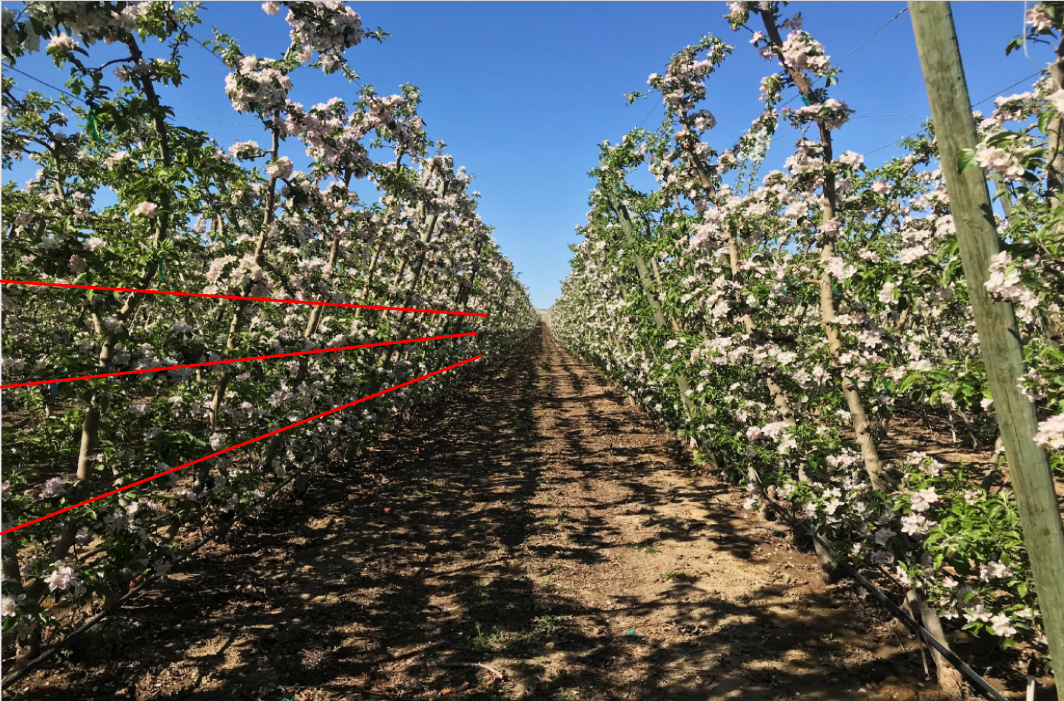}
    \caption{V-trellised commercial apple orchard in Prosser, WA used for field evaluation of the robotic system. Red lines highlight the section of the canopy where the integrated system was tested.}
    \label{fig_chap6:thinningtrelisses}
\end{figure}

\begin{figure}[!h]
    \centering
    \includegraphics[scale=0.38]{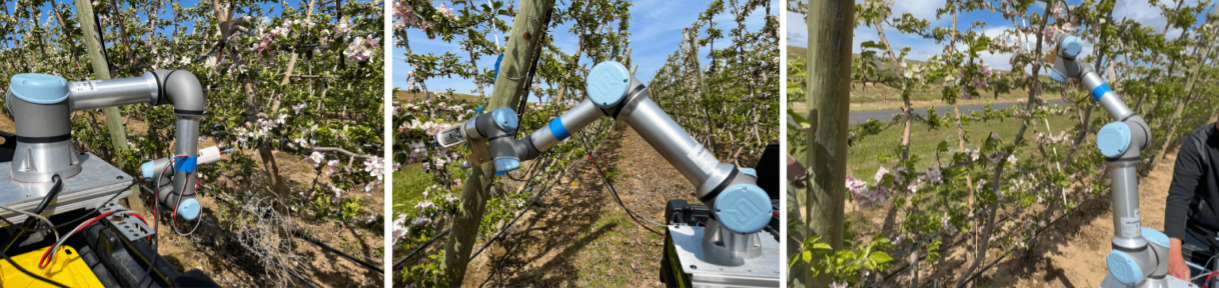}
    \caption{Examples images depicting the integrated system in action during field evaluation in a commercial apple orchard around three trellis wires: Lower trellis wire (Left), Center trellis wire (Middle), Upper trellis wire (Right)}
    \label{fig_chap6:robotinaction}
\end{figure}

The end effector approach direction for both boundary and center thinning was orthogonal to the cluster surface. In this field evaluation, a total of 148 flower clusters were attempted for thinning (Boundary Thinning:72, Center Thinning: 76). Out of 148 samples, three samples were discarded since the robot started unpredictable or erratic behavior. Since the robotic manipulator has a reachability of 850 mm, the thinning experiments were performed in the clusters located on the branches trained to three trellis wires (second, third, and fourth from the bottom) of the 7-wire apple tree canopies that ensured good reachability of the manipulator  (see Figure \ref{fig_chap6:thinningtrelisses}). Figure \ref{fig_chap6:robotinaction} shows the images captured during the thinning process covering different sections of the tree canopy.

\section{Software Architecture} \label{sec_chap6:SoftwareArchitecture}
\begin{figure}[!h]
\centering
\includegraphics[scale=0.7]{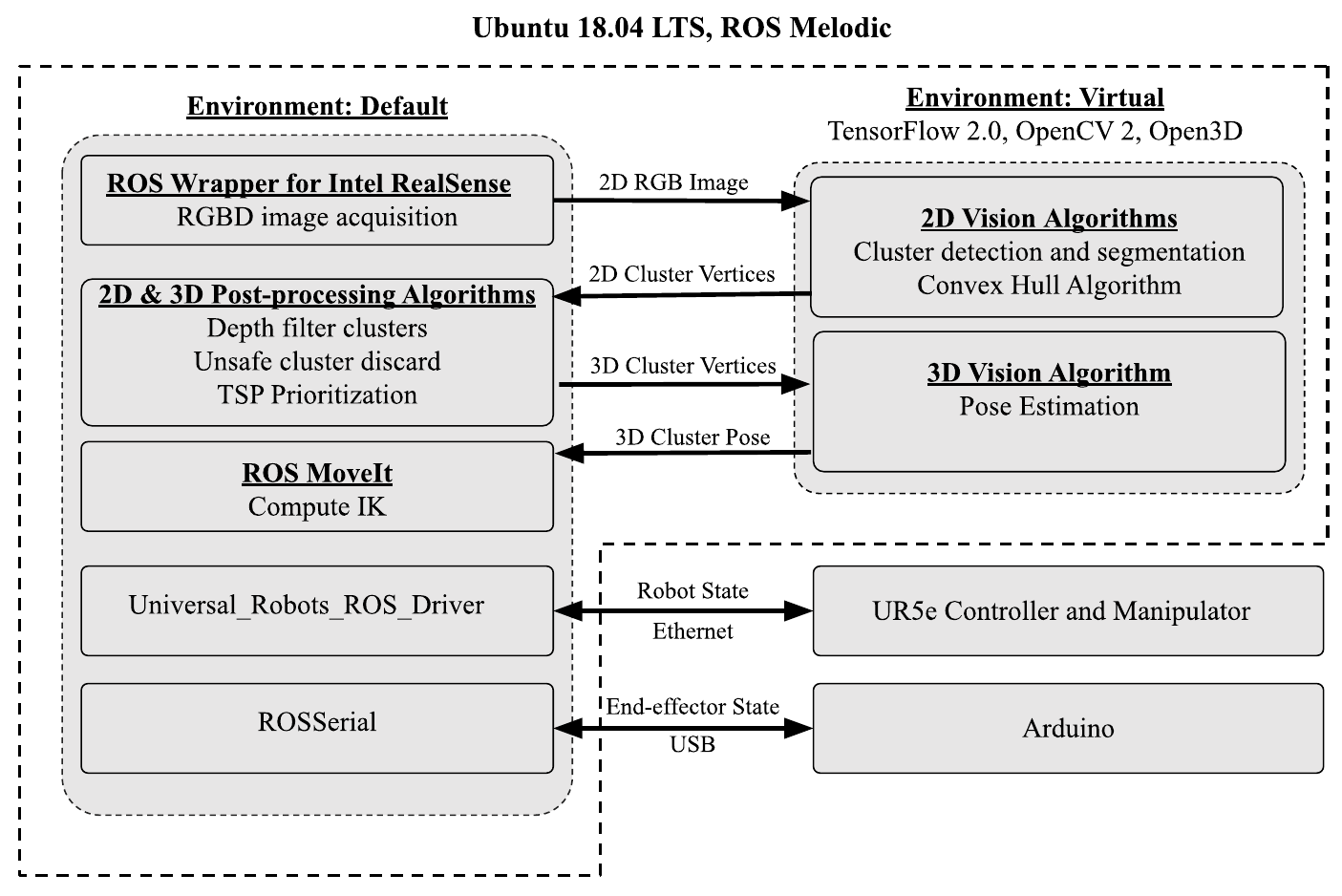}
\caption{Flowchart of the software architecture used in this study. The diagram illustrates information flow among different components of the robotic thinning system.}
\label{fig_chap6:SoftwareArchitecture}
\end{figure}

Figure \ref{fig_chap6:SoftwareArchitecture} details the software architecture of the integrated system used in this study. All the software components were encapsulated and executed using a Dell Alienware15R4 laptop within ROS Melodic. However, two distinct ROS workspaces and computational environments were set up: Default, Virtual. The Virtual environment was responsible for executing the majority of 2D and 3D machine vision algorithms and operations, such as 2D cluster detection and segmentation and 3D pose estimation. The Default environment was the core of the system, which communicated and managed the peripheral devices such as the IntelRealSense camera, UR5e controller and manipulator, Arduino end-effector controller, and the Virtual environment. Communication among different components in the software layer was managed using the ROS ecosystem.

At the beginning of an experiment, the IntelRealSense D435i camera acquired an RGBD image of the canopy. As discussed in section \ref{sec_chap6:MachineVisionSystem}, the 2D RGB image was used as the input to the instance segmentation algorithm, which detected and estimated the cluster boundary. The estimated cluster boundary was simplified using the Convex Hull algorithm.

Given a set of 3D positions of the feasible/accessible clusters, and the starting position of the manipulator, the next step was to determine the shortest possible route such that the end-effector visits each cluster exactly once and returns to the start position with minimal travel cost (Euclidean distance). This problem was modeled as the classic optimization problem known as the Traveling Salesman Problem (TSP), which is an “NP-hard” problem because of the difficulty in finding solutions for large instances. The TSP was solved by employing Local Search algorithm with OR-Tools, an open-source software to solve optimization problems \citep{googleortools}). The Local Search algorithm started with a random end-effector path that visits each cluster and iteratively improves the path by swapping the locations of a pair of clusters and checking if the new path was better until no further improvements could be made. Based on the solution from solving the TSP, the cluster visit sequence for the end effector was prioritized, and the thinning was subsequently performed accordingly.

\begin{figure}[!hb]
    \centering
    \includegraphics[scale=0.9]{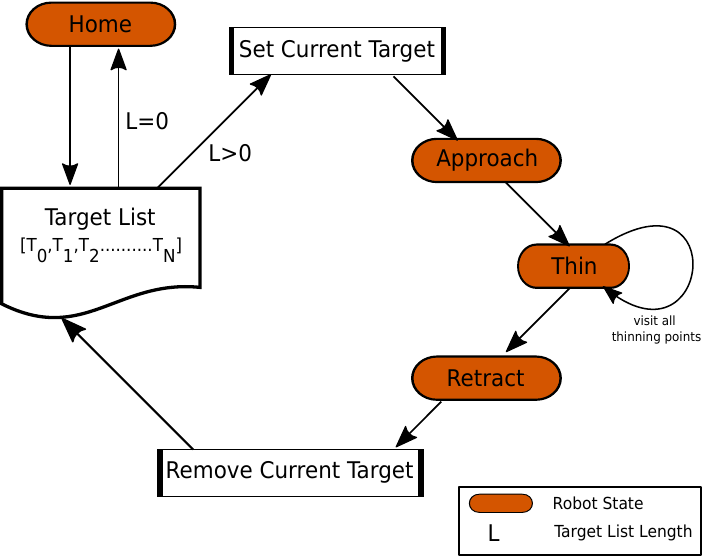}
    \caption{Simplified illustration of the thinning process as a state machine. This state machine was executed after the target flower clusters were finalized and prioritized.}
    \label{fig_chap6: robot state machine}
\end{figure}

The thinning process of clusters were modeled as a state machine, as illustrated in Figure \ref{fig_chap6: robot state machine}.  During each thinning cycle, the system was in one of the four states: \textit{Home}, \textit{Approach}, \textit{Thin}, or \textit{Retract}. At the beginning and the end of the experiment, the system was in an idle \textit{Home} state, waiting for commands. During the initial \textit{Home} state of the robot, the system carried out various tasks including image acquisition, cluster detection and segmentation, pose estimation, as well as post-processing procedures such as filtering out unreachable clusters, implementing Convex Hull-based boundary simplification, and cluster visit prioritization. 

Let $[T_0,T_1,T_2.........T_N]$ be $N$ target clusters selected and prioritized for thinning. During the thinning operation, the end-effector was navigated to three vantage points: \textit{Approach Point}, \textit{Thinning Start Point}, and \textit{Retract Point}. 

The system states \textit{Approach}, \textit{Thin}, and \textit{Retract} were defined for motion planning and navigation of the end-effector to these three vantage points. The \textit{Approach Point} was chosen at a 10 cm offset distance from the cluster centroid along the cluster surface normal. The \textit{Approach Point} was introduced as an intermediate target such that the end-effector was aligned orthogonally to the cluster surface and ready to approach the cluster. In this study, the canopy was assumed a 2D planar structure. Orienting the end-effector along the cluster surface normal at the \textit{Approach Point} before physically interacting with the cluster required only a simplified linear motion from the \textit{Approach Point} to the \textit{Thinning Start Point}. The \textit{Thinning Start Point} was the first point within the cluster where the end-effector physically interacted with the flowers in the target cluster. For center thinning, the \textit{Thinning Start Point} was the cluster centroid, while for Boundary Thinning, the \textit{Thinning Start Point} was the first vertex in the polygonal boundary.

The thinning process started after the end-effector reached the \textit{Thinning Start Point}, and the system entered to the \textit{Thin} state. The system would be in the \textit{Thin} state for a 2-second for center thinning or until all the boundary points were visited for boundary thinning. The \textit{Retract Point} was same as the \textit{Approach Point}, except it was visited after the thinning process was completed. \textit{Approach Point} and \textit{Retract Point} ensured the system’s safety by forcing majority of the motion planning tasks to occur outside of the canopy, thereby reducing the risk of collision with canopy parts such as branches, trunks, and trellis wire. Once thinning process was successfully completed or failed for a target cluster, the cluster was removed from the Target List, and the process was repeated until all clusters were visited. The system returned to the \textit{Home} state at the end of the thinning cycle.

% \newpage

% \bibliographystyle{apalike}
% \bibliography{jfrExampleRefs}

%% file: sections/Results.tex
\section{Results} \label{sec_chap6:Results}
\subsection{Machine Vision System}
\begin{figure}[!hb]
    
    \centering
    \begin{minipage}{0.48\textwidth}
        \centering
        \subfloat[Segmented Clusters]{\label{field_chap6:a}\includegraphics[scale=0.175]{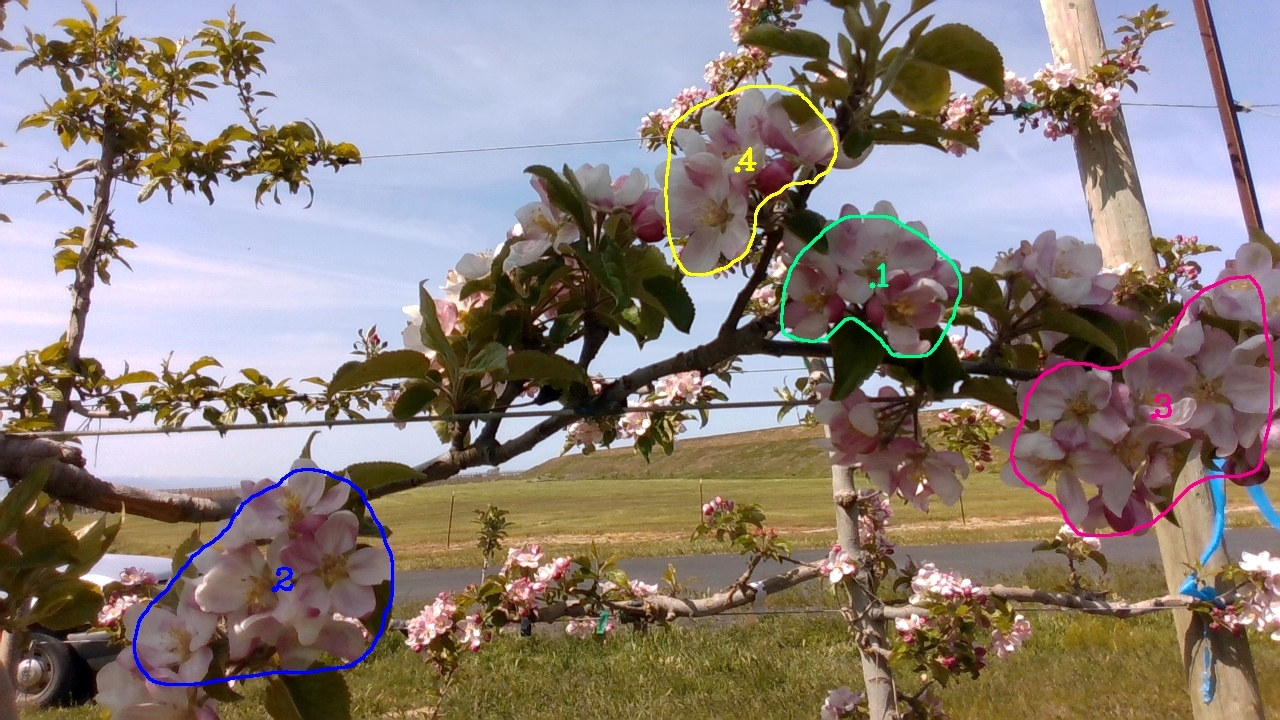}}
    \end{minipage}
    \begin{minipage}{0.48\textwidth}
        \centering
        \subfloat[Clusters after Convex Hull boundary simplification]{\label{field_chap6:b}\includegraphics[scale=0.175]{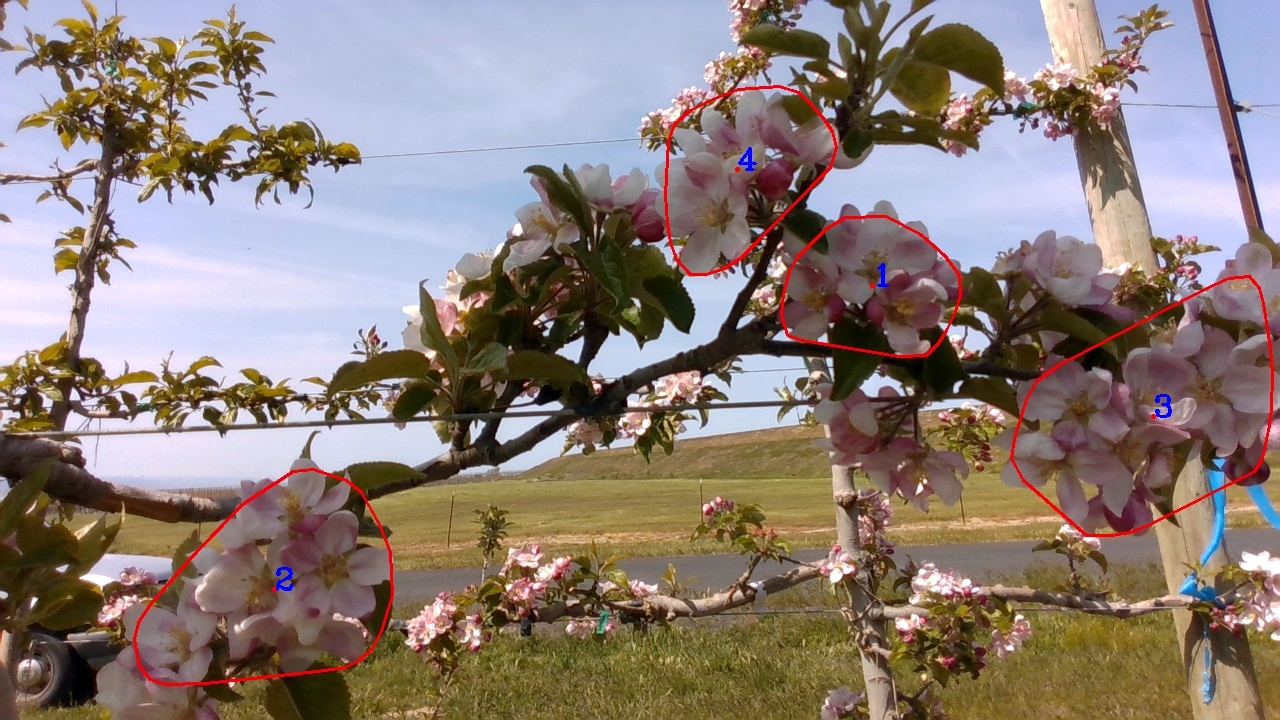}}
    \end{minipage}

    \begin{minipage}{0.48\textwidth}
        \centering
        \subfloat[Segmented Clusters]{\label{field_chap6:c}\includegraphics[scale=0.175]{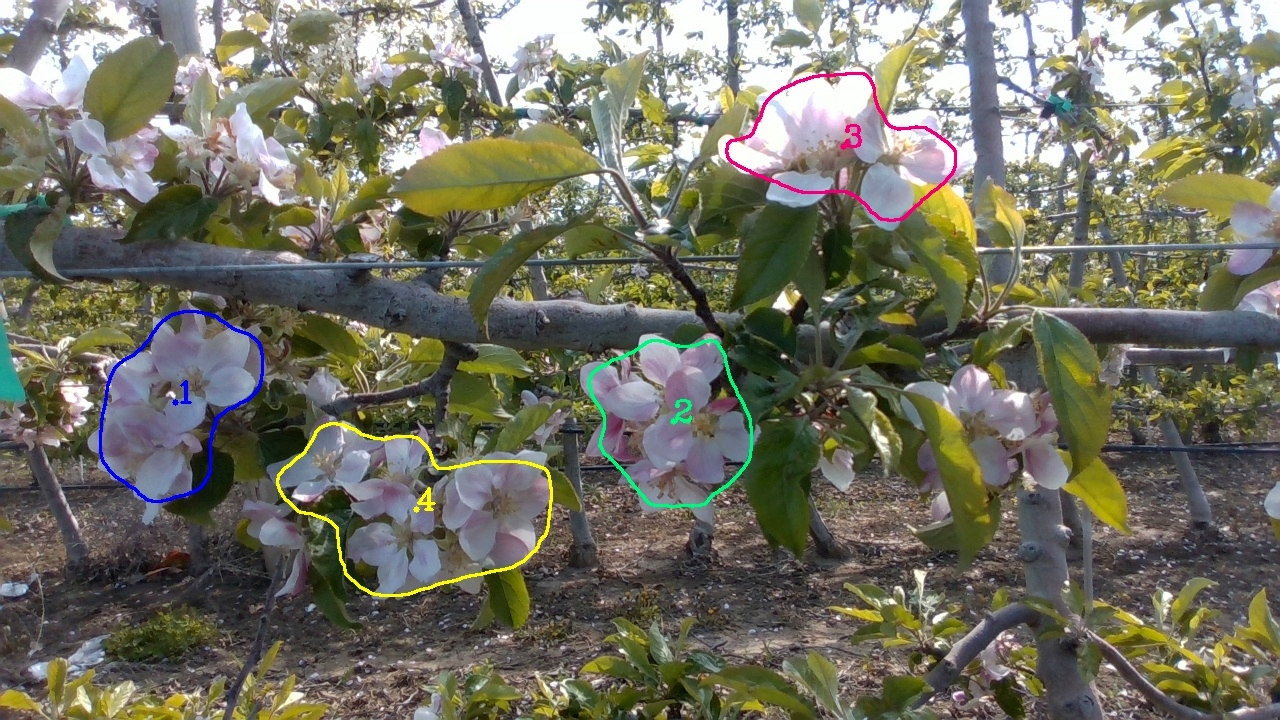}}
    \end{minipage}
    \begin{minipage}{0.48\textwidth}
        \centering
        \subfloat[Clusters after Convex Hull boundary simplification]{\label{field_chap6:d}\includegraphics[scale=0.175]{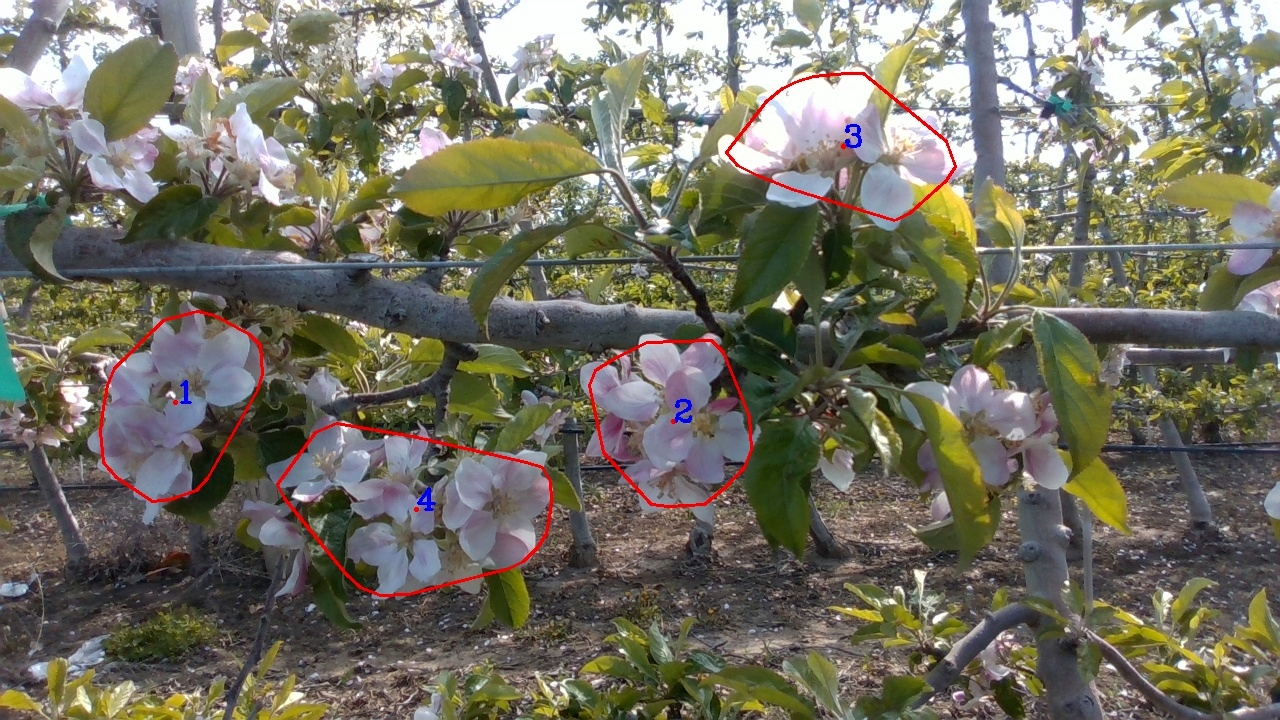}}
    \end{minipage}
    
    \caption{Results of the machine vision system in identifying and delineating cluster boundaries in the varying background, lighting condition and experiment time [a) Early stage full bloom, morning c) Late stage full bloom, afternoon]. b) and c) represent the resulting images for a) and c), respectively after the boundaries were simplified using Convex-Hull algorithm.}
    \label{fig_chap6:maskrcnn_outdoor_good}
\end{figure}

Figure \ref{fig_chap6:maskrcnn_outdoor_good} shows the example results (qualitative) of the Mask R-CNN-based machine vision algorithm for cluster detection and segmentation during the field tests. Image in Figure 10a is a representative image during the early stage (notice the pink color shades) of bloom whereas Figure 10c is the image from the same field two days later with a different background. Apple flowering window was brief, spanning only a few days from full bloom to petal fall. This resulted in rapid changes in flower appearance, along with additional orchard variability such as variable lighting conditions, background, and wind.  Despite the limited training data, biological variability in the canopy, and the outdoor dynamic variability in orchard, the Mask R-CNN algorithm demonstrated promising results in identifying the flower clusters. Quantitative evaluation in the test dataset showed that the Mask R-CNN algorithm achieved a mean Average Precision (mAP) of 0.89 to segment cluster boundaries with True Positive Rate (True Positive/(True Positive + False Positive) of  98\% in identifying the clusters.

As shown in Figure \ref{field_chap6:a} and \ref{field_chap6:c}, the detected flower cluster boundaries were approximated by polygonal structures with a large number of vertices that were used to represent the cluster boundary smoothly and accurately. Figure \ref{field_chap6:b} and \ref{field_chap6:d}show the corresponding results of the Convex Hull algorithm. The algorithm substantially reduced the number of polygonal vertices while approximating the cluster boundary with desirable smoothness. The use of the Convex Hull algorithm was expected to decrease the system cycle time and reduce computational complexity without substantially compromising the system’s thinning performance.

\begin{figure}[!hb]
    \centering
    \begin{minipage}{0.48\textwidth}
        \centering
        \subfloat[Segmented Clusters]{\label{fieldbad_chap6:a}\includegraphics[scale=0.175]{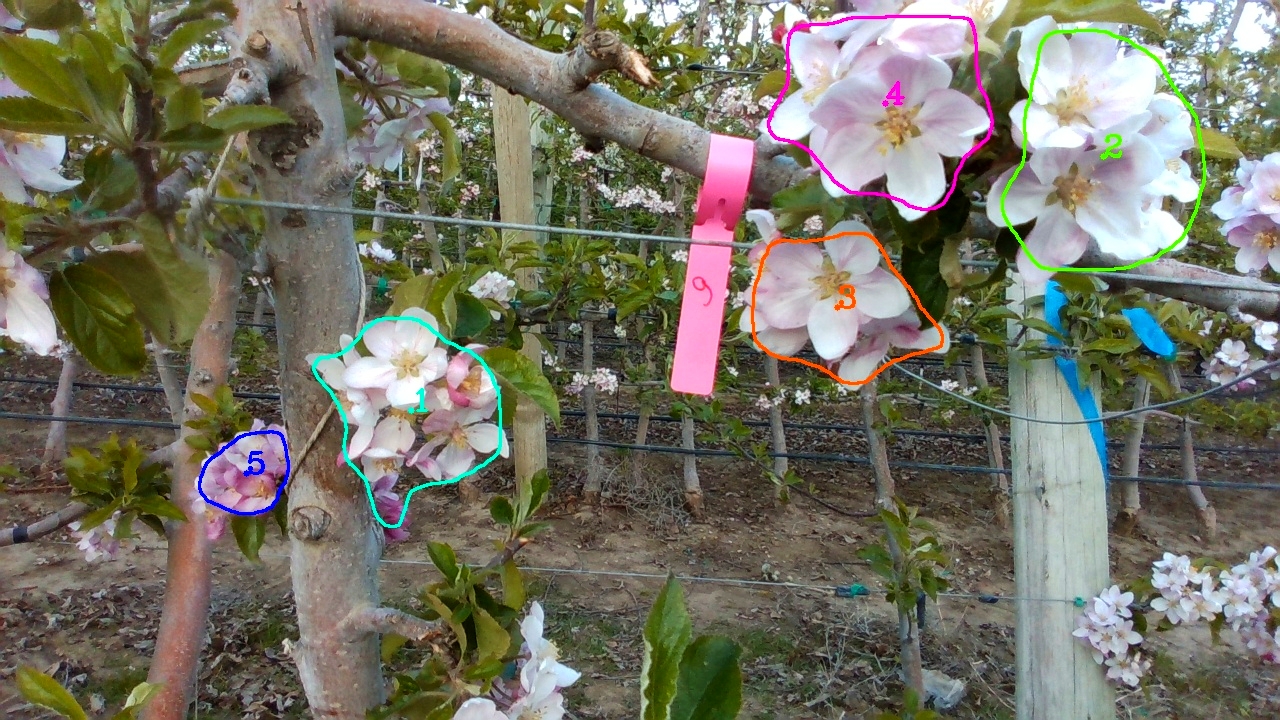}}
    \end{minipage}
    \begin{minipage}{0.48\textwidth}
        \centering
        \subfloat[Filtered clusters]{\label{fieldbad_chap6:b}\includegraphics[scale=0.175]{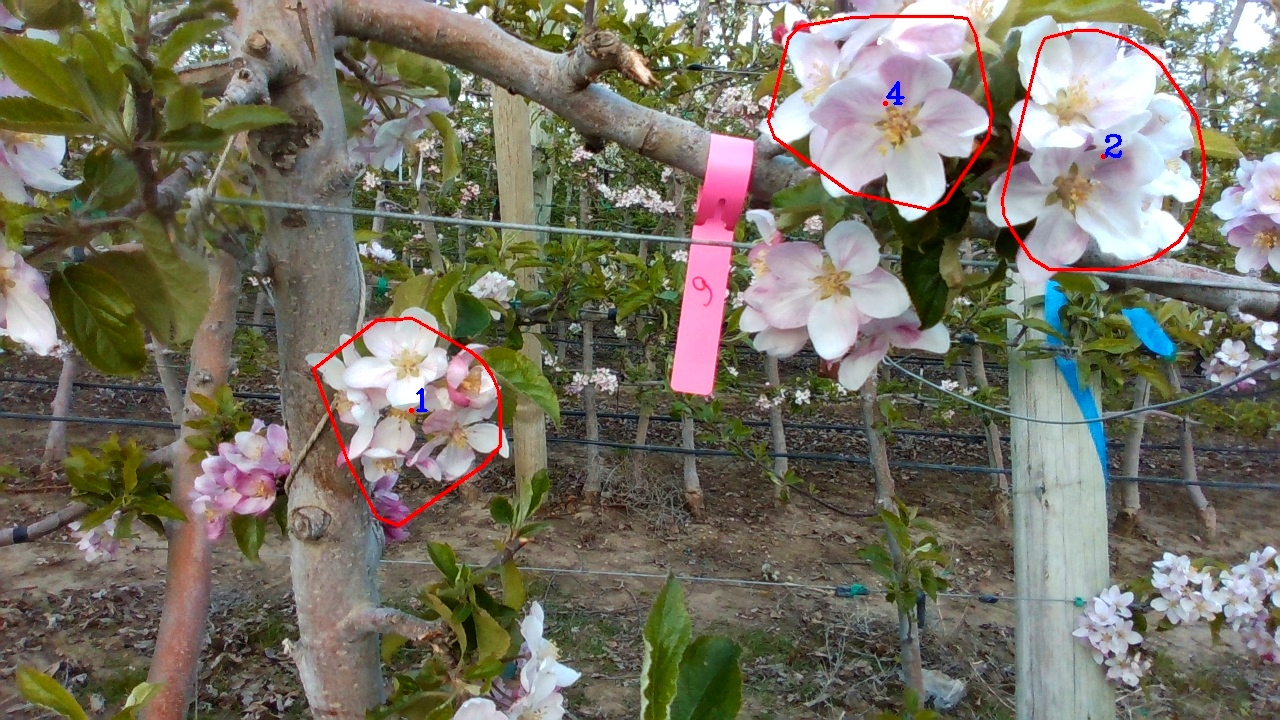}}
    \end{minipage}
    
    \caption{Each segmented flower clusters were evaluated for safe reachability for the robotic manipulator a) Segmented flower clusters from the machine vision algorithm b) Finalized flower clusters after post-processing and removal of cluster potential to collide robotic manipulator with trunk (Cluster ID: 5) and trellis wire (Cluster ID: 3) during thinning.}
    \label{fig_chap6:maskrcnn_outdoor_bad}
\end{figure}

As discussed in the earlier sections, the segmented clusters were automatically further examined for reachability, followed by manual removal of clusters in the unsafe locations for robot to access. Figure \ref{fig_chap6:maskrcnn_outdoor_bad} shows an example of clusters that were removed from the target clusters for thinning because of their proximity to the trunk and trellis wire. Table \ref{tab_chap6:flowers_human removed} shows that $\sim 84\%$ and $\sim 91\%$ of the segmented clusters were accepted for boundary and center thinning tests. It is also worth noting that the operator rejected a higher proportion ($\sim 13\%$ vs. $\sim 5\%$) of clusters for boundary thinning compared to center thinning. For boundary thinning, the operator should consider the potential collision between the robot and the canopy parts at all the vertices in the cluster boundary. Assessment of the cluster centroid for potential collision was generally enough for center thinning approach.

\begin{table}[!ht]
    \centering
    \caption{Detected flower clusters vs clusters attempted to thin}
    \label{tab_chap6:flowers_human removed}
    \setlength{\tabcolsep}{3pt}
    \scalebox{1}{
    \begin{tabular}{llcccc}
        \hline \hline
        &Clusters & Boundary Thinning & Center Thinning &Total \\
        \hline
        &Accepted &201 &161 &362 \\
        & Rejected (Automatic) &8 &7 &15 \\
        & Rejected (Human in loop) &30 &8 &38 \\ 
        \hline
        &Total Detected &239 &176 & 415 \\
         \hline \hline
    \end{tabular}
    }

\end{table}

\subsection{Evaluation of Motion Planning}
Once the candidate clusters were identified, the cluster poses (position and orientation) were estimated and used as the input for the robot motion planner. A motion planning success rate $\sim$ 93\%  was achieved for boundary thinning and $\sim$ 89\% center thinning.  Table \ref{tab_chap6:motionplanning} shows the number of successful and failed attempts in determining robot trajectory using the TRAC -IK Inverse Kinematics (IK) solver. Majority of failure cases in motion planning was not finding the IK solution. In a few cases, although IK solution was found, the manipulator started to follow dangerous motion (Non-optimal IK), and the tests were aborted.

Once the candidate clusters were identified, the cluster poses (position and orientation) were estimated and used as the input for the robot motion planner. Table \ref{tab_chap6:motionplanning} details the motion planning statistics. In the majority of motion plan failure cases Inverse Kinematics (IK) solution was not found. While in some of the attempts, although the IK solution was found, the manipulator started to follow dangerous motion (Non-optimal IK), and such experiments were aborted. We achieved a motion planning success rate of $\sim$ 93\% for boundary thinning and $\sim$ 89\% center thinning.   
\begin{table}[!ht]
    \centering
    \caption{Clusters accepted as thinning candidate vs successful attempts}
    \label{tab_chap6:motionplanning}
    \setlength{\tabcolsep}{3pt}
    \scalebox{1}{
    \begin{tabular}{llcccc}
        \hline \hline
         &Motion Plan &Boundary Thinning &Center Thinning &Total\\
         \hline
         &No IK   &11 &15 &26 \\
         & Non-optimal IK &3 &2 &5\\
         & Success &187 &144 &331 \\
         \hline
         & Accepted Clusters &201 &161 &362\\
         \hline \hline
    \end{tabular}
    }
\end{table}

\subsection{Thinning Efficiency}
Out of the 415 clusters detected by the machine vision system, thinning was successfully conducted in 331 ($\sim 80 \%$) clusters (Boundary Thinning: 187 and Center Thinning: 144). As discussed earlier, the flower thinning in orchards requires the removal of a certain proportion of flowers while leaving another certain proportion intact. However, out of the 331 flower clusters, 27 clusters were completely removed during thinning resulting in $92\%$ success in conducting proportional flower thinning. The boundary thinning approach successfully thinned $67.2\%$of the flowers, while the center thinning approach thinned $59.5\%$of the flowers.

Although it was desirable to completely remove excess flowers and keep the remaining ones intact, achieving such an outcome was highly difficult to achieve. Clusters grow randomly in the canopy leading to highly variable and uncertain position and orientation of flowers within each cluster. Additionally, the end-effector interacted with the flowers in different ways while physically engaged with them. As a result, the thinned flowers were categorized into four major categories:

\begin{itemize}
    \item \textit{Completely Removed}: The flower was completely removed; only the remaining part was the flower stem. 
    \item \textit{Petal and Anther Removed}: The petals and anthers were completely removed, and the stigma and/or pollen tube were damaged. Anthers are easily detachable from the flowers compared to the stigma and pollen tube, which is directly connected to the ovary
    \item \textit{Petal Removed}: More than $80\%$ of petal removed and minor effect in anther and no effect in stigma and pollen tube.
    \item \textit{Saved}: More than $80\%$ of petals saved and no effect on anther, stigma, and pollen tube.
\end{itemize}

\begin{figure}[h!]
\centering
\includegraphics[width=0.48\textwidth]{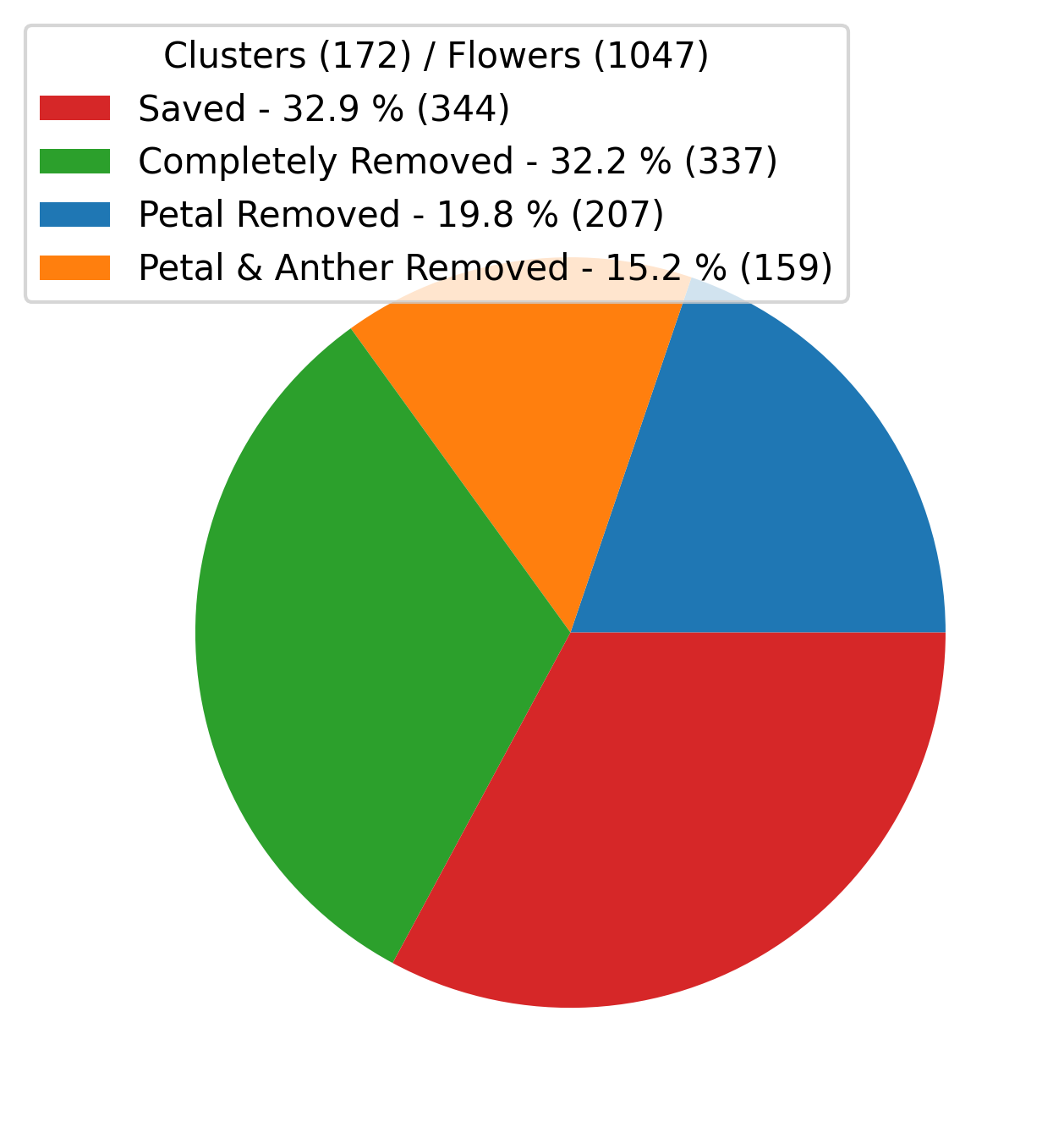}
\includegraphics[width=0.48\textwidth]{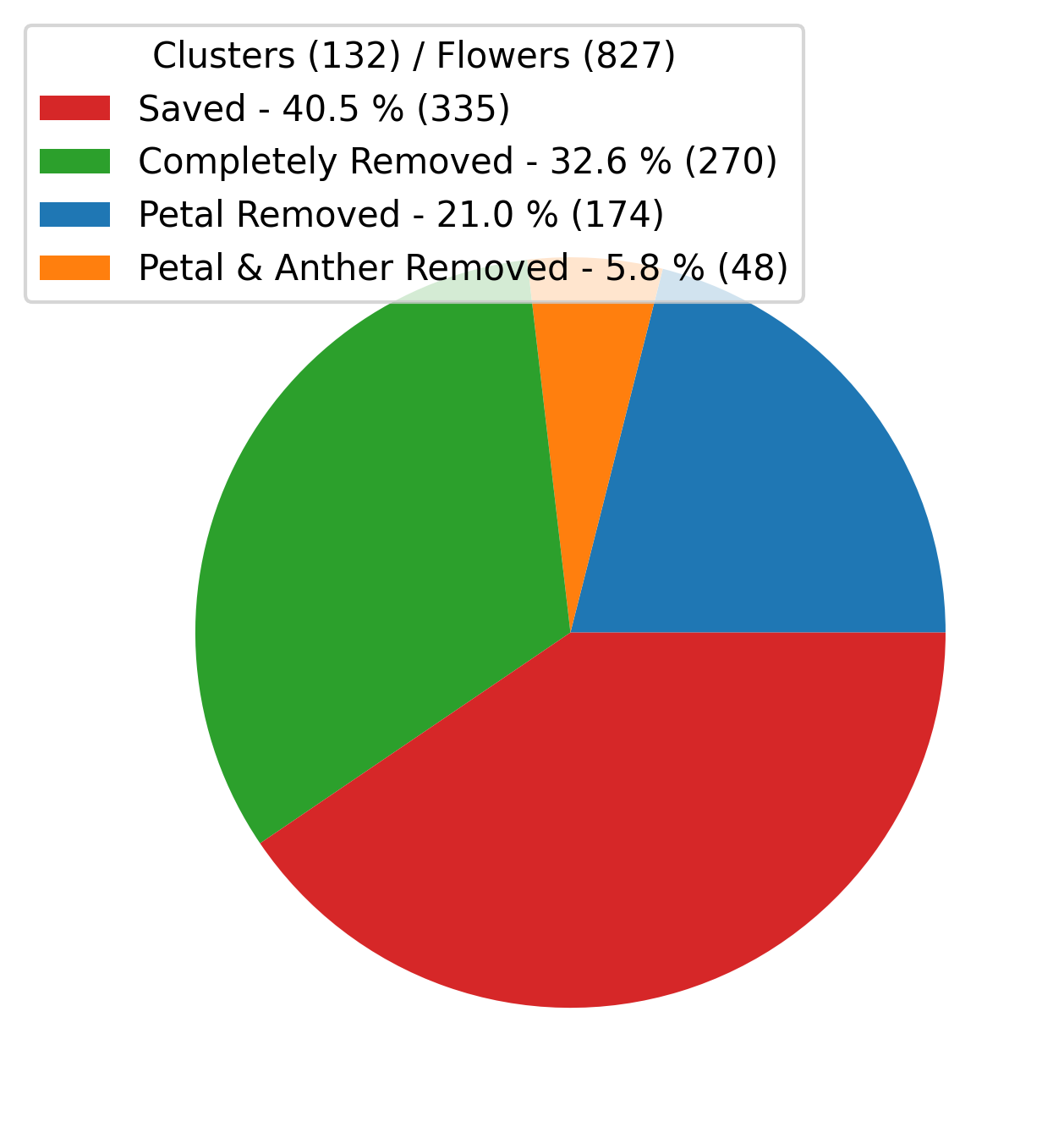}

\caption{Thinning efficiency breakdown of boundary thinning (left) and center thinning (right) approach. The boundary thinning approach thinned larger proportion of flowers compared to center thinning since the end effector was actuated around the cluster boundary during the boundary thinning operation..}
\label{_chap6ThinningEfficiency}

\end{figure}

Flower petals serve the primary function of attracting pollinators, such as bees, with their vibrant and colorful appearance. Removal of the petals can reduce the chances of pollinators visiting and pollinating the flower. The stigma and pollen tubes are essential for transporting pollen grains to the ovary for fertilization. If the stigma and pollen tubes are damaged, the flower may not be able to receive pollen, or pollens may not be able to reach the ovules, preventing fertilization. Figure \ref{_chap6ThinningEfficiency} illustrates the thinning breakdown of thinning efficiency both boundary and center thinning. Both boundary and center thinning approaches completely removed $ \sim 33\%$ of the flowers. However, the center thinning method was more successful in saving flowers ($40.5\%$) compared to the boundary thinning method ($32.9\%$). This was expected since the boundary thinning method moves the end-effector around the cluster boundary, while the center thinning method only targets the center of the clusters. As mentioned earlier, some flowers encountered various levels of damage during the thinning process, including the removal of petals and anthers and damage to the stigma and pollen tube. These damages significantly reduced the flowers’ ability to attract pollinators and perform pollination and pollen germination. Therefore, to evaluate the thinning efficiency, all the flowers that were not completely saved were considered. 
These results indicated that the thinning intensity can be adjusted by modifying the thinning strategy. The proposed robotic blossom thinning system showed the ability to perform variable rate thinning, which is crucial for selective blossom thinning that could be optimizable for achieving desired crop-load in individual trees based on specific field conditions.

\subsection{Cycle Time Evaluation}
\begin{figure}[!hb]
\centering
\includegraphics[width=0.48\textwidth]{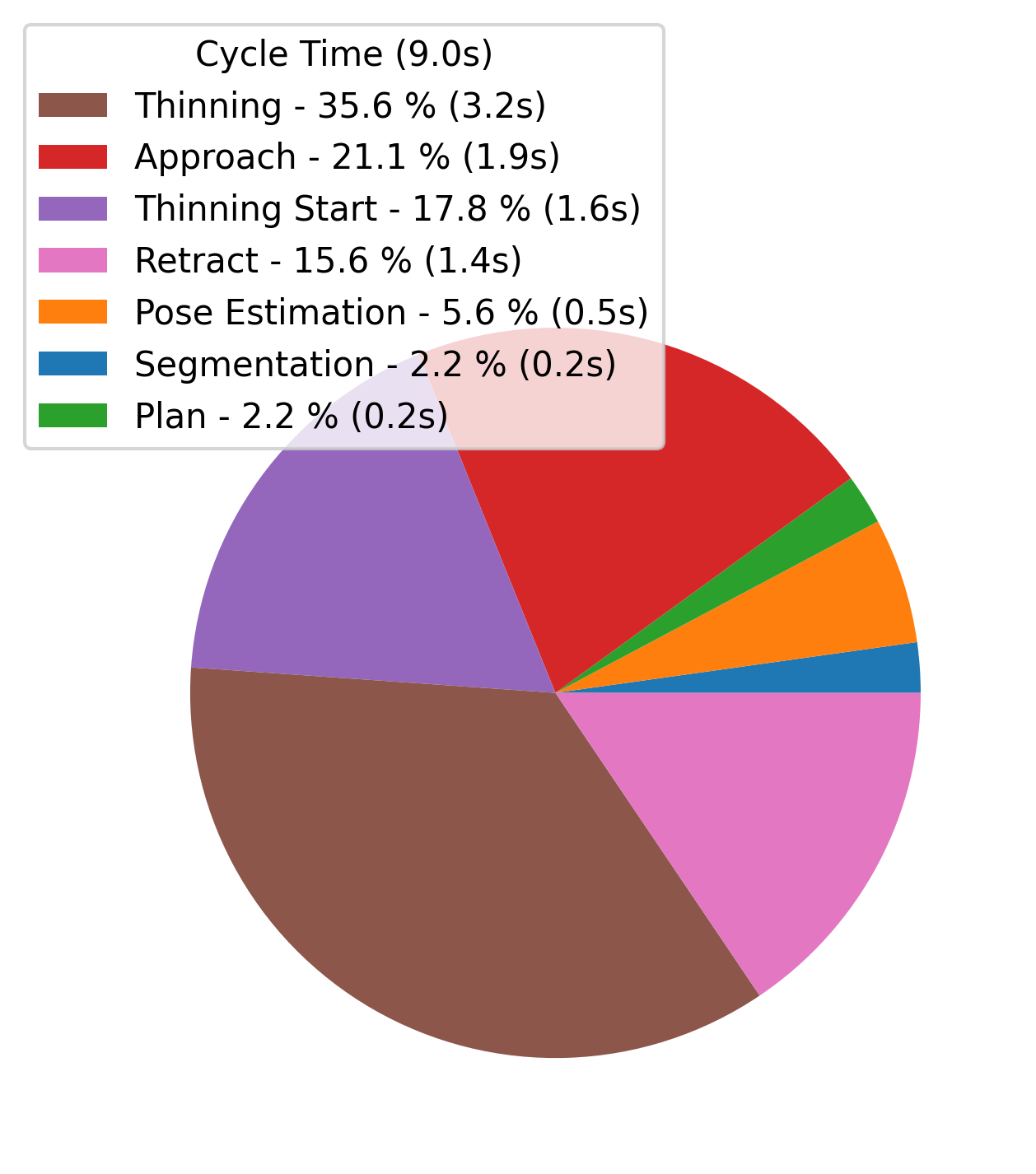}
\includegraphics[width=0.48\textwidth]{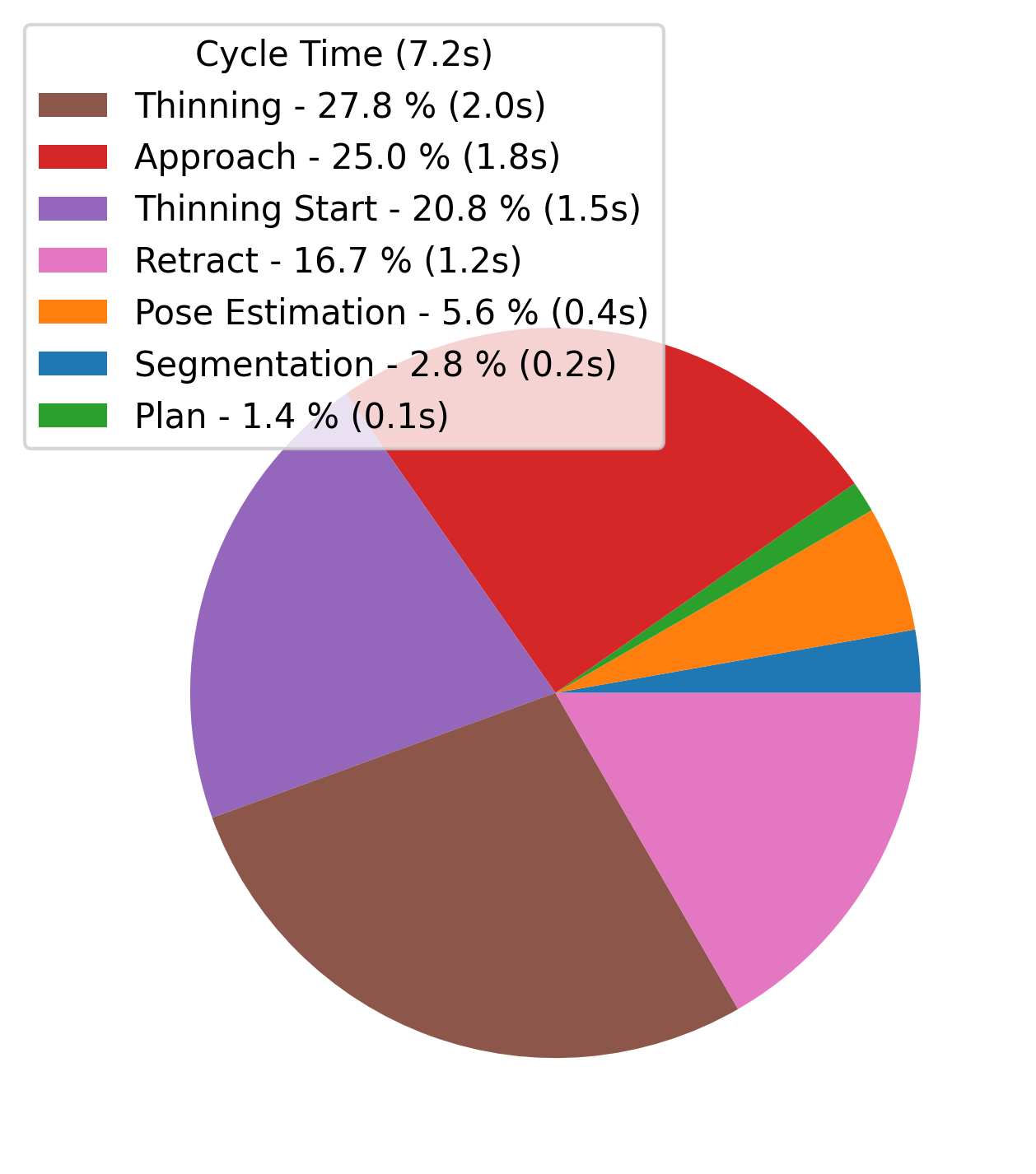}

\caption{Cycle time breakdown for boundary thinning (left) and center thinning (right) approaches evaluated for robotic flower thinning in a commercial orchard, Prosser. WA. While boundary thinning approach thinned larger proportion of flowers compared to center thinning, boundary thinning led to increased cycle time compared to center thinning..}
\label{fig_chap6:CycleTime}

\end{figure}

The cycle time for each cluster for boundary thinning and center thinning was 9.0 seconds and 7.2 seconds, respectively as shown in Figure \ref{fig_chap6:CycleTime}. The time required for all operations in center and boundary thinning was relatively similar with the exception of \textit{Thinning} time. The \textit{Thinning} time for boundary thinning was 1.2 seconds $(60\%)$ more compared to the center thinning. For boundary thinning, the thinning time varied based on the cluster size, while it was constant at 2 seconds for center thinning regardless of cluster size. As the overall thinning process constituted a number of sub-tasks including the machine vision system, manipulator and motion planning system, and the end-effector system, the time required for each sub-task was recorded and the overall cycle time was evaluated (see Figure \ref{fig_chap6:CycleTime}). The subtasks were categorized as \textit{Segmentation, Motion Planning, Pose Estimation, Cluster Approach, Thinning,} and \textit{Retraction tasks}. The \textit{Segmentation} time was referred to as the time taken for image acquisition and object detection and segmentation. The overall \textit{Motion Planning} time for the robotic manipulator was the \textit{Plan time}, while \textit{Pose Estimation} time was the time required to estimate the cluster pose in 3D space. The \textit{Approach} time was the execution time from the robot’s \textit{Home} position to the \textit{Approach} position of the cluster or from the \textit{Retract} position of the previously visited cluster to the \textit{Approach} position of the current cluster. \textit{Thinning Start} time was the execution time from the \textit{Approach} position to the \textit{Thinning Start} position. For boundary thinning, the \textit{Thinning Start} position was the first vertex of the cluster boundary polygon, while for center thinning, it was the cluster centroid. \textit{Retract} time referred to the execution time from the \textit{Thinning Start} position to the \textit{Approach} position. Based on the thinning efficiency results presented in the previous sub-section and the cycle time, it was evident that the thinning strategy can influence both thinning efficiency and cycle time. It was found that the boundary thinning offered a slightly higher efficiency in the cost of increased cycle time compared to the same achieved with the center thinning approach.

% \newpage
% \bibliographystyle{apalike}
% \bibliography{jfrExampleRefs}

%% file: sections/Conclusion.tex
\section{Discussions, Lessons Learned, and Future Direction} \label{sec_chap6:Discussion}
In this work, a robotic flower-thinning system was designed, developed, integrated, and evaluated in a commercial apple orchard. Based on the results and qualitative observations during the development and evaluation of this system, the following insights are presented.
\begin{itemize}

    \item Robotic blossom thinning, similar to other field operations such as harvesting and pruning, is a complex and challenging operation for both humans and robotic systems. Additionally, the thinning season offers limited window of operation to be effective. Flowering time varies based on the weather and temperature, and it can last a couple of days up to a week, while harvesting and pruning seasons last for a couple of months. However, the thinning window could get expanded (may be not as much as for harvesting) because of the varieties and locations of planting though some varieties at one specific location may offer a narrow window of operation/opportunity.

    \item Deep learning algorithm showed robust performance in outdoor environment. However, in some instances, the deep learning algorithms identified clusters in the background canopies as target clusters. This was anticipated to some degree since the background canopy in the V-trellised training architecture (the one used in this study) was closer to the foreground canopy because of the inverted pyramid structure. However, this issue may be not critical in the case of a vertical fruiting wall architecture, where the background canopies are separated with a considerable distance. In the instances when background clusters were detected in the current study, the depth filtering algorithm filtered them out. Nevertheless, the system could benefit from illumination-invariant active lighting imaging technologies for added system robustness against variable lighting conditions, which would also then facilitate nighttime operation.

    \item The use of the commercial-grade camera helped reduce the cost of the machine vision system. The use of commercial camera also helped in software development, prototyping, and troubleshooting because of the readily available camera software drivers, ROS camera interface, and community support.

    \item The employed single-view imaging helped reduce the thinning cycle time with the scan-and-go operation. However, the single-view system could not offer more complete and accurate information about the object of interest during the field trial. In some instances, even if clusters were separated in 3D space, they appeared to overlap when projected onto a 2D image, leading the segmentation algorithm to mistakenly assume two clusters for a single cluster. Furthermore, the cluster pose estimation algorithm was based on the point cloud obtained from single-view imaging might not be sufficient to estimate the correct cluster pose. The system could benefit from multi-view imaging, which could be accomplished by moving the local camera to various waypoints or by incorporating one or more global cameras to capture the surroundings more comprehensively.

    \item The use of two thinning strategies, boundary thinning and center thinning, helped achieve varying proportion of flower removal from the cluster. 
    While boundary thinning strategy removed higher proportion of flowers compared to center thinning it came with the cost of increased cycle time.  Since the end effector had to visit the cluster boundaries compared to center thinning, the thinning time varied depending upon the cluster size, and was often greater than the fixed 2 seconds thinning time for center thinning. The thinning strategies can be adjusted based on factors such as flowering intensity, cluster spacing, cultivar, canopy crop-load-bearing capacity, and thinning speed to achieve desired crop load. For instance, if the cluster size is large for center thinning, the end-effector could be actuated in multiple locations. 

    \item The use of the human in the loop system was crucial in ensuring the safe reachability of the manipulator without collision with other canopy parts such as trunk, branches and trellis wire. Few instances were observed where the system was in collision and the test was aborted. However, the current human in the loop system also added additional workload to the operator. The current machine vision systems could be improved such that obstacles such as branches, trellis wires could be identified and locate. Based on the obstacle pose, the thinning and collision avoidance strategy could be changed automatically reducing the workload to the operator. For instance, center thinning could be employed instead of boundary thinning or by opting not to thin a particular cluster at all based on the severity of the safety concern.

    \item The planar tree canopy architecture allowed flexible working space for the robotic manipulator. Even with the limited reachability of 850mm of the UR5e manipulator, most of the flowers were easily accessible for thinning. However, new trellis-trained canopy architectures also came with challenges posed by the presence of trellis wires and support posts. Trellis wires are thin, which could be detected by deep learning algorithms, but it could be challenging to estimate depth and avoid colliding with them while approaching a target flower cluster. Additionally, accessing flowers in the back surface of the V-trellis canopy system was challenging. Unlike the vertical fruiting wall architecture, where the canopy would have been from both sides, V-trellised architecture had closed spaces between the adjacent canopies making the canopy accessible from only one side.

    \item The current system was designed to compute the cluster pose. Estimation of the cluster pose helped precisely orient the end-effector along the normal to the cluster surface. A future study direction could be to investigate the significance of cluster pose estimation. Analysis of various approach angles to the flower cluster removal could help reduce the system cycle time and improve thinning efficiency in the future. Such a study could also help reduce the overall system complexity and then cost using linear actuators with fewer degrees of freedom compared to UR5e manipulators (such as the one used in this study).

    \item The current system was an open-loop system which allowed the system to operate at higher speed resulting in reduced cycle time. A closed-loop system with visual serving would enable the system to make decisions based on the information acquired momentarily. However, utilizing a closed-loop system could increase the cycle time. Hence study on the trade-off between the benefits and drawbacks of open and closed-loop systems could be carried out in the future.
    
    \item Although field evaluation experiment was carried out thoroughly, the sample size was relatively small. The technology developed requires bigger-scale, multi-year field tests for further validation.
    
    \item Human-in-the-loop systems and human-robot collaboration are crucial techniques to improve the efficiency and autonomy of various agricultural robotic. Machines may fail, and the generated motion plans could be nonoptimal, primarily when the machine operates in an unstructured environment common in agriculture.  Hence, the humans-in-the-loop can be used to help with such a situation and make some critical decisions.

    % \item Currently, there is a limited publicly accessible agricultural dataset, particularly for apple flower detection. Machine learning models require a large amount of training dataset for improved accuracy and robustness. A large number of datasets with varying example scenarios could help improve the robustness of the machine vision system for flower detection against dynamic background, biological variability, and vital differences between crop cultivars.

\end{itemize}

\section{Conclusions}\label{sec_chap6:Conclusion}
Blossom thinning is one of the crucial field operations that require a large workforce for narrow bloom season to thin down flowers selectively as machine or chemical thinning solutions lack the desired selectivity, control and effectiveness. In this study, a robotic flower thinning system was designed, developed, integrated, and tested for selective flower thinning in a commercial orchard. The system leveraged stereo vision, artificial intelligence, and robotic manipulator to solve one of the complex field operations in tree fruit production. Based on the results from this study, the following conclusions were drawn.

\begin{itemize}
    \item Deep learning-based machine vision algorithm efficiently learned to detect and segment flower clusters in the foreground canopy while successfully discarding the flowers in the trees in adjacent rows and the background sky, both of which had similar appearances as flowers. Out of all identified clusters using the machine vision system, $98\%$ of identified clusters were correct with mean average precision of 0.89 to estimate the cluster boundary. 

    \item Field evaluation of the integrated system in commercial orchard showed promising potential of targeted, and selective robotic blossom thinning in commercial orchards. The integrated robotic blossom thinning system was able to effectively remove varying proportions of blossoms depending on the thinning strategy: boundary, center. The boundary thinning approach thinned higher proportion of the flowers $(67.2\%)$ with higher cycle time of 9.0 seconds per flower cluster compared to the center thinning approach which thinned $57.4\%$ of flowers with cycle time of 7.2 seconds per flower clusters. 
\end{itemize}

The proposed system has several prospects for future advancements and improvements. The implementation of the proposed system could facilitate the development of commercially viable robotic blossom thinning systems aiding the adoptability of robotic systems intended for operation in tree fruit crops.